%% file: active_learning_main_modular.tex
\documentclass[conference]{IEEEtran}
\IEEEoverridecommandlockouts
\usepackage[utf8]{inputenc}
\usepackage{tikz}
\usepackage{float}
\usepackage{forest}
\usetikzlibrary{arrows.meta, shapes.geometric, positioning, fit, backgrounds}
\usepackage{booktabs}
\usepackage{multirow}

\usepackage{cite}

\usepackage{amsmath,amssymb,amsfonts}\usepackage{dsfont}
\usepackage{algorithm}
\usepackage{algpseudocode}

\usepackage{graphicx}
\usepackage{textcomp}
\usepackage{xcolor}

\usepackage{caption}
\input{library/math_commands}

\def\BibTeX{{\rm B\kern-.05em{\sc i\kern-.025em b}\kern-.08em
    T\kern-.1667em\lower.7ex\hbox{E}\kern-.125emX}}
\begin{document}

\title{Active Learning Methods for Efficient Data Utilization and Model Performance Enhancement
\thanks{\hspace*{-\parindent}\rule{3.8cm}{0.4pt} \\ 
$\dagger$: Corresponding author: Junhao Song (junhao.song23@imperial.ac.uk).}
}

\author{
\IEEEauthorblockN{Chiung-Yi Tseng}
\IEEEauthorblockA{\textit{AI Agent Lab}\\
Berkeley, USA\\
ctseng.p@gmail.com}
\and
\IEEEauthorblockN{Junhao Song$\dagger$}
\IEEEauthorblockA{\textit{Imperial College London}\\
London, United Kingdom\\
junhao.song23@imperial.ac.uk}
\and
\IEEEauthorblockN{Ziqian Bi}
\IEEEauthorblockA{\textit{Purdue University}\\
West Lafayette, USA\\
bi32@purdue.edu}
\and
\IEEEauthorblockN{Tianyang Wang}
\IEEEauthorblockA{\textit{University of Liverpool}\\
Liverpool, United Kingdom\\
tianyangwang0305@gmail.com}
\and
\IEEEauthorblockN{Chia Xin Liang}
\IEEEauthorblockA{\textit{JTB Technology Corp.}\\
Kaohsiung, Taiwan\\
marcus.chia@ai-agent-lab.com}
\and
\IEEEauthorblockN{Xinyuan Song}
\IEEEauthorblockA{\textit{Emory University}\\
Georgia, USA\\
xsong30@emory.edu}
\and
\IEEEauthorblockN{Ming Liu}
\IEEEauthorblockA{\textit{Purdue University}\\
West Lafayette, USA\\
liu3183@purdue.edu}
}

\maketitle

\begin{abstract}
In the era of data-driven intelligence, the paradox of data abundance and annotation scarcity has emerged as a critical bottleneck in the advancement of machine learning. This paper gives a detailed overview of Active Learning (AL), which is a strategy in machine learning that helps models achieve better performance using fewer labeled examples. It introduces the basic concepts of AL and discusses how it is used in various fields such as computer vision, natural language processing, transfer learning, and real-world applications. The paper focuses on important research topics such as uncertainty estimation, handling of class imbalance, domain adaptation, fairness, and the creation of strong evaluation metrics and benchmarks. It also shows that learning methods inspired by humans and guided by questions can improve data efficiency and help models learn more effectively. In addition, this paper talks about current challenges in the field, including the need to rebuild trust, ensure reproducibility, and deal with inconsistent methodologies. It points out that AL often gives better results than passive learning, especially when good evaluation measures are used. This work aims to be useful for both researchers and practitioners by providing key insights and proposing directions for future progress in active learning.
\end{abstract}


\input{diagrams/active_learning_structure.tex}

\section{Introduction}

\input{diagrams/TiDAL}

\input{diagrams/CuttingPlain}

\input{sections_modular/01_02_new_introduction_section}

\section{Methodologies}

\input{sections_modular/07_uncertainty_estimation_and_sampling_strategies_in_active_learning.tex}

\input{tables/BALD}

\input{sections_modular/08_deep_learningbased_active_learning_methods.tex}

\input{sections_modular/13_interactive_and_questiondriven_learning_in_active_learning}

\section{Challenges}

\input{sections_modular/05_fairness_and_bias_in_active_learning.tex}

\input{diagrams/ALBench}

\input{diagrams/AdaptiveSupervision_ObjectDetection}

\input{sections_modular/09_class_balance_and_imbalance_issues_in_active_learning.tex}

\section{Evaluations}

\input{sections_modular/10_evaluation_metrics_and_benchmarks_for_active_learning.tex}

\section{Applications}

\input{sections_modular/03_applications_of_active_learning_in_computer_vision.tex}

\input{algorithms/GAL/GAL}

\input{algorithms/GAL/Aquisition_function}

\input{sections_modular/04_active_learning_for_natural_language_processing_tasks.tex}

\input{algorithms/Margatina_2022/Margatina_2022}

\input{diagrams/AADA}
\input{diagrams/Empowering}
\input{sections_modular/06_domain_adaptation_and_transfer_learning_in_active_learning.tex}

\input{diagrams/DEAL}

\input{sections_modular/11_realworld_applications_and_case_studies_of_active_learning.tex}

\input{sections_modular/14_active_learning_for_specialized_tasks.tex}

\section{Conclusion}

\input{sections_modular/16_conclusion.tex}

\bibliographystyle{ieeetr}
\bibliography{cite}

\end{document}

%% file: library/math_commands.tex
\def\eqref#1{equation~\ref{#1}}

\def\1{\bm{1}}

\DeclareMathAlphabet{\mathsfit}{\encodingdefault}{\sfdefault}{m}{sl}
\SetMathAlphabet{\mathsfit}{bold}{\encodingdefault}{\sfdefault}{bx}{n}

\DeclareMathOperator*{\argmin}{arg\,min}

%% file: diagrams/active_learning_structure.tex
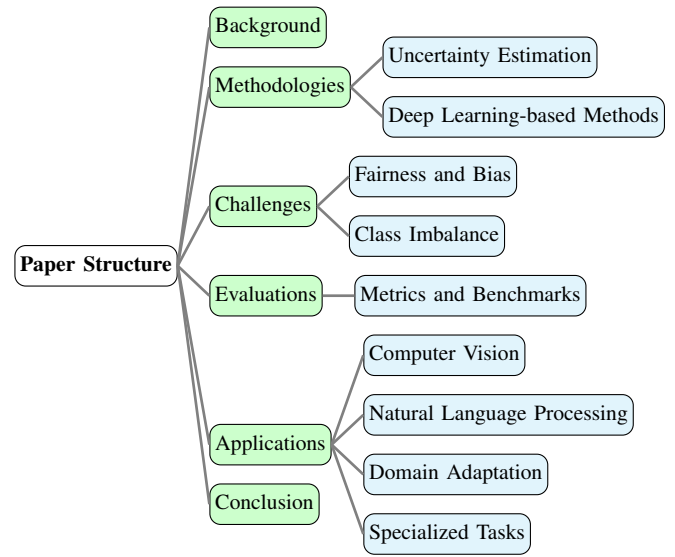
\begin{figure}[htbp]
  \centering
  \resizebox{\columnwidth}{!}{
    \begin{forest}
      for tree={
        grow=east,
        draw,
        rounded corners,
        minimum height=1.5em,
        anchor=base west,
        child anchor=west,
        parent anchor=east,
        base=left,
        rectangle,
        align=left,
        edge+={gray, line width=1pt},
        inner sep=2pt,
        font=\footnotesize,
      },
      where level=1{fill=green!20}{},
      where level=2{fill=cyan!10}{},
      where level=3{fill=purple!10}{},
      where level=4{fill=orange!20}{},
      where level=5{fill=gray!10}{},
      [\textbf{Paper Structure}
        [Conclusion]
        [Applications
          [Specialized Tasks]
          [Domain Adaptation]
          [Natural Language Processing]
          [Computer Vision]
        ]
        [Evaluations
          [Metrics and Benchmarks]
        ]
        [Challenges
          [Class Imbalance]
          [Fairness and Bias]
        ]
        [Methodologies
          [Deep Learning-based Methods]
          [Uncertainty Estimation]
        ]
        [Background]
      ]
    \end{forest}
  }
  \caption{Structure of this Active Learning (AL) survey.}
\end{figure}

%% file: diagrams/TiDAL.tex
\begin{figure*}
\includegraphics[width=0.9\linewidth]{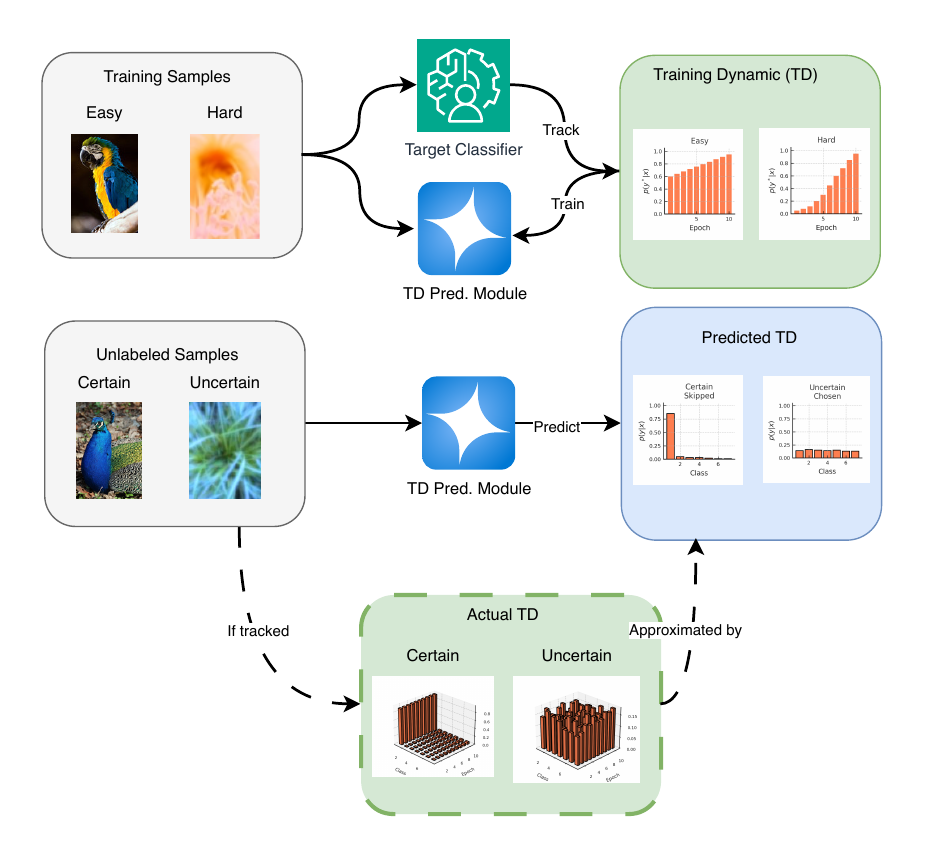}
\centering
\caption{The TiDAL framework builds on the insight that the training dynamics (TD) of samples can vary, even when their final predicted probabilities $p(y^*|x)$ are identical (upper row). This motivates the use of intermediate training signals by leveraging TD as a source of rich information. To address the computational burden of tracking TD across large-scale unlabeled data, the framework employs a prediction module to estimate TD, rather than explicitly recording it for every sample (lower row).}
\end{figure*}

%% file: diagrams/CuttingPlain.tex
\begin{figure*}[t]
\includegraphics[width=0.9\linewidth]{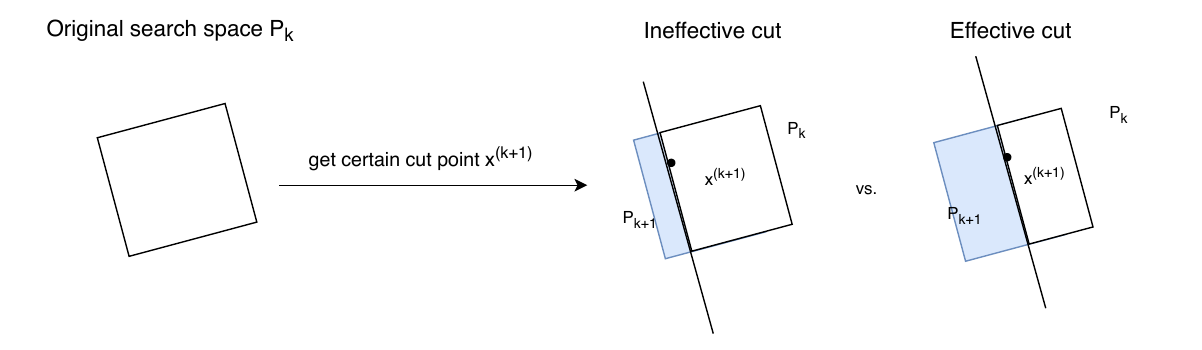}
\centering
\caption{Illustration of general cutting-plane method on single iteration.}
\end{figure*}

%% file: sections_modular/01_02_new_introduction_section.tex
Active Learning (AL) is a key technique in contemporary machine learning. It minimizes labeling costs while maintaining model quality or enhancing model accuracy \cite{10.1007/978-3-642-41398-8_16, ding2023active}. Contrary to conventional supervised learning where there are randomly labeled examples used, AL selects the most informative unlabeled instances for labeling. This improves the training and makes it scalable \cite{bosser2021model, abraham2020rebuilding}. AL is particularly useful in domains where obtaining labeled examples is difficult or costly, such as medical imaging, scientific simulations, and low-resource languages.

Interest in AL has increased with the rise of deep learning. Many new techniques are being created for Deep Neural Networks (DNN) \cite{bosser2021model, zhang2022survey, jukic-snajder-2023-smooth}. The techniques employ uncertainty, differences in features, and learned representations for selecting useful samples. Research also indicates that improved benchmarks and clear communication are necessary to see when AL succeeds or fails \cite{10.1007/978-3-642-41398-8_16, abraham2020rebuilding}. Simulations demonstrate that the performance of AL will vary significantly with task, model, and data \cite{10.1007/978-3-642-41398-8_16, abraham2020rebuilding}.

In principle, researchers have applied models such as Bayesian approaches \cite{kruschke2008bayesian} and Markov decision processes \cite{puterman1990markov} to describe how AL chooses data over time \cite{2306.08001}. New metrics nowadays enable comparison of the strategies adopted by AL in addition to accuracy. Examples include uncertainty, calibration, and robustness measures \cite{abraham2020rebuilding, jiang2024actively}. Using these tools, the construction of more trustworthy and interpretable AL systems becomes possible. 

AL focus has transitioned. Previous methodologies predominantly evaluated the model. Today, more attention is on the data. Model-centered methodologies leverage model uncertainty or confidence for data selection \cite{bosser2021model, 2201.03947}. Data-centered methodologies consider the structure and diversity of the data. They employ techniques such as pseudolabeling, curriculum learning, and synthetically generated data \cite{chu2016can, 2304.05246}. The two perspectives in combination provide a more complete description of learning between model and data. 

Despite advances, there are still challenges. One of them is that AL results do not necessarily replicate between tasks. Benchmarks may vary in how they normalize data, define label boundaries, or calculate measures. To address that, the community requires common standards and open-source tools \cite{jukic-snajder-2023-smooth, jiang2024actively}. Another challenge is integrating AL into continual learning, where tasks and data vary over time. AL is also applied in sophisticated fields such as quantum computing and physics-based learning, where the data is expensive but of high quality \cite{ding2023active, 2302.14567}. The uses across fields demonstrate that science will be facilitated by AL. Some of the methods of AL also model human-like behavior such as attention, curiosity, and self-regulation \cite{2011.03733, takezoe2023deep}. All these concepts serve towards making more intelligent and more versatile learning systems.

This paper presents an overview of the main theories, techniques, and outstanding challenges of AL. It weaves concepts from cognitive science, deep learning, and statistics together with real-world systems. We aim to describe central concepts, compare techniques, indicate how testing of AL occurs, and identify what remains to be resolved. This should be of most value to new researchers as well as experienced researchers.

%% file: sections_modular/07_uncertainty_estimation_and_sampling_strategies_in_active_learning.tex
\textbf{Uncertainty Estimation and Sampling Strategies:} estimation of uncertainty and sampling strategies are key aspects of AL, which facilitate the identification of informative examples for model improvement with minimal labeling. New uncertainty estimation techniques have been proposed in recent works, with evaluation of the effectiveness of prior techniques and derivation of the theoretical underpinnings of uncertainty-based AL \cite{li2024adaptive}. For example, epistemic and aleatoric uncertainty are distinguished within AL, with the argument that using epistemic uncertainty for selecting informative examples is a better characterization \cite{nguyen2019epistemic}. An examination of model mismatch has shown that uncertainty-based AL underperforms random sampling when model capacity is limited \cite{2408.13690}. Different acquisition functions have been proposed for tackling the problem. Further, non-asymptotic convergence of uncertainty sampling has been proved for binary and multi-class classification for noise-free and noisy cases \cite{raj2022convergence}.

Deep Bayesian methods for AL have shown stronger performance on NLP tasks \cite{carvalho2024deep}. Siddhant and Lipton \cite{1808.05697} demonstrate that Bayesian AL by disagreement, using uncertainty estimates from Dropout or Bayes‑by‑Backprop \cite{hernandez2015probabilistic}, surpasses uncertainty sampling and i.i.d baselines \cite{he2021towards}. A belief-function‑based uncertainty method of sampling \cite{hoarau2024evidential} reduces computational effort and captures label uncertainty by design. The advances reflect increasing attention on Bayesian and evidential approaches for active learning. All researches stress the necessity of uncertainty estimation for active learning. They investigate uncertainty sampling, Bayesian active learning, and evidential uncertainty sampling. They also mention the effect of model capacity and mismatch on performance.

Convergence analysis and bounds have gained prominent focus. Lu et al. \cite{2306.08954} re-benchmarked pool-based active learning for binary classification. Mussmann and Liang \cite{mussmann2018relationship} explored the relationship between error and data efficiency for uncertainty sampling. Other research suggests other approaches like weighted uncertainty sampling \cite{1909.04928} and learning the model uncertainty \cite{kye2023tidal}. A stopping criteria using deterministic generalization bounds has been proposed to automatically stop active learning \cite{2005.07402}. All these contrasting opinions reflect the controversy in the field.

However, model capacity and mismatch are still important challenges \cite{2408.13690}. Computational intensity is also an issue. Simplifications of the computational task have become available \cite{hoarau2024evidential}. There are limited theoretical guarantees as well. There remains a need for more work on the principles and limitations of active learning.

In summary, recent research has significantly improved our knowledge of uncertainty estimation and sampling techniques in AL. The creation of novel uncertainty estimation techniques, model mismatch analysis, and the provision of theoretical bounds have all improved our understanding in this regard. Nonetheless, ongoing challenges and limitations serve as evidence that there remains room for further research in AL studies, specifically towards addressing model capacity complexities, computational complexities, and theoretical underpinnings. As research in AL proceeds, it will likely birth new advances, drawing from the work of these recent studies \cite{hoarau2024evidential, 2408.13690, raj2022convergence}.

%% file: tables/BALD.tex
\newcommand{\std}[1]{ \normalfont \color{darkgray}\footnotesize{$\pm$#1} }

\begin{table*}[ht]
  \vspace{2em}
  \centering
  \resizebox{0.8\textwidth}{!}{
\begin{tabular}{l|cc|cc}
\toprule
& \multicolumn{2}{c|}{\bf Sensitive attribute} & \multicolumn{2}{c}{\bf Minority group} \\
\midrule
\bf @ 10\%  & Predictive parity & Accuracy \% & Predictive parity & Accuracy \% \\

\midrule
    \bf Uniform &  10.73 \std{2.70} &  87.23 \std{1.77} &  4.67 \std{0.76} &  88.53 \std{1.66} \\
    \bf Uniform + GRAD $\lambda=0.5$ &   7.58 \std{2.16} &  87.37 \std{0.98} &  2.03 \std{1.46} &  84.98 \std{0.88} \\
    \bf Uniform + GRAD $\lambda=1$ &   5.50 \std{1.51} &  83.69 \std{2.79} &  1.38 \std{0.00} &  86.52 \std{1.81} \\
    \bf AL-Bald &   3.56 \std{1.70} &  91.66 \std{0.36} &  2.11 \std{0.19} &  \textbf{94.08 \std{0.10}} \\
    \bf AL-Bald + GRAD $\lambda=0.5$ &   2.16 \std{1.13} &  \textbf{92.34 \std{0.26}} &  \textbf{0.74 \std{0.15}} &  93.38 \std{0.84} \\
    \bf AL-Bald + GRAD $\lambda=1$ &   \textbf{1.27 \std{0.88}} &  90.31 \std{0.84} &  0.75 \std{0.65} &  91.72 \std{0.67} \\
    \midrule
    \bf Balanced Uniform(Oracle) &  10.34 \std{1.97} &  86.91 \std{1.83} &  2.40 \std{0.65} &  90.66 \std{0.41} \\
    \bf Balanced Uniform(Oracle) GRAD $\lambda=0.5$ &   5.15 \std{1.39} &  88.40 \std{2.02} &  1.88 \std{0.01} &  90.24 \std{0.70} \\
    \midrule
    \bf REPAIR~\cite{Li_2019_CVPR} & 0.54 \std{0.11} & 94.52 \std{0.19} &  1.06 \std{0.44} & 92.84 \std{0.33} \\
\bottomrule
\end{tabular}}
  \caption{Comparison between BALD and uniform labeling after 10\% of the dataset has been labeled. $\lambda$ is the weight of the gradient reversal term. Both metrics are evaluated on a balanced held-out set. Standard deviation is reported by repeating the experiment with 3 different random seeds \cite{2104.06879}.}
  \label{tab:result}
  \vspace{-0.5em}
\end{table*}

%% file: sections_modular/08_deep_learningbased_active_learning_methods.tex
\textbf{Deep Learning-based Active Learning Methods:} deep learning-based AL garnered considerable attention over the past years \cite{lewis1994heterogeneous,lewis1995sequential,luo2013latent,settles2008analysis,vijayanarasimhan2014large,iglesias2011combining}. However, numerous approaches do not work and are even inferior to random sampling. Entropy-based AL remains a solid baseline and usually outperforms single-model methods, and the pretraining options, initial budget, and budget step size can have unexpected effects on results \cite{2403.14800}.

Some research uses cutting-plane algorithms for deep networks. These are a decent substitute for gradient-based alternatives \cite{2410.02145}. AL also employs pseudo-labels over real labels. Incremental and cumulative learning assist with enhanced performance \cite{bosser2021model}. Evaluation of AL for real-world tasks such as semi-supervised learning and object detection is crucially important \cite{2403.14800, 1912.05361}. Li et al. also identify that we still have no comprehensive survey that explains deep AL methodologies, their benefits and drawbacks, as well as applications \cite{li2024survey}.

Deep AL is extremely sensitive to the training conditions, like data augmentation \cite{1912.05361}. It can enhance performance if coupled with semi-supervised learning or gradient-free optimization \cite{2410.02145, bosser2021model}. A study reveals model-centric (model-centered and query-strategy-centered) and data-centric (pseudo-labels and training modes-centered) aspects of active learning with deep neural network models, e.g., training mode, query strategies, availability of unlabeled data, initial training, and network specifications \cite{bosser2021model}. Despite the sophisticated alternatives, there remains controversy regarding whether simple entropic approaches outperform alternatives \cite{2403.14800, li2024survey}.

Technical aspects also play a part. The application of ReLU networks and network depth influences AL performance in image classification and segmentation tasks \cite{2410.02145}. It’s also necessary to test approaches on both real and synthetic datasets in order to see what their limits are \cite{2410.02145, bosser2021model}. Stating the statistical properties of chosen examples and test error estimation also serves to estimate deep AL \cite{bosser2021model}.

There has been improvement, but there are still challenges. We have not yet had a full comparison of various deep AL techniques \cite{li2024survey}. The outcomes depend heavily on training and experiments \cite{1912.05361}. There exists a great demand for AL techniques that perform effectively with big datasets and few labels \cite{2403.14800, 2410.02145}. Even more, approaching the design of AL techniques that achieve the optimal solution reliably remains a major challenge \cite{2410.02145}.

%% file: sections_modular/13_interactive_and_questiondriven_learning_in_active_learning.tex
\textbf{Interactive and Question-Driven Learning in Active Learning:} this is an important part of active learning. It helps machines ask questions and learn through dialogue. For example, the INTERACT framework lets large language models join in student-teacher conversations. This leads to up to 25\% better performance \cite{2412.11388}. It shows that interaction is valuable—machines learn better when they ask and receive information step by step. Another method, structural query-by-committee \cite{1803.06586}, gives a general algorithm for interactive learning tasks. It works well even when there is noise in the data.

The difference between passive and active learning is a common topic. Active learning usually gives better results \cite{2412.11388, 2011.03733}. Some tools compare different strategies, helping users see their pros and cons \cite{2111.04936}. Other studies compare human-like learning with machine-based systems like INTERACT \cite{2412.11388, 2011.03733}.

Many papers show that human-like active learning, interactive, question-based learning performs better than passive learning in both human and machine studies \cite{2011.03733}. This kind of learning also focuses on explainability. Visualization tools help people understand how models learn and make decisions \cite{2111.04936}. Some papers also connect active learning with curiosity. They suggest that active learning is like curiosity-driven exploration \cite{1312.2936, 2011.03733}.

On the technical side, several methods are used. These include student-teacher dialogues \cite{2412.11388}, query-by-committee \cite{1803.06586}, and knowledge distillation \cite{2011.03733}. Tools like the visual interface in \cite{2111.04936} give useful views into how training happens. Still, there are problems. Scalability is one issue \cite{2412.11388}. Dealing with noise and uncertainty is another \cite{1803.06586}. Also, better evaluation metrics are needed \cite{2111.04936}. Making humans and machines work together is also a challenge \cite{2011.03733}.

Interactive active learning can be applied in natural language processing \cite{zhang2022survey}, visual question answering \cite{1711.01732}, and education \cite{2406.13903}. But to make more progress, we need to fix its current problems, e.g., building better metrics and connecting human-style and machine-style learning. In the future, interactive learning may change how machines learn—making them faster, more accurate, and easier to understand \cite{misra2018learning, 2306.08001, 2202.00254}.

%% file: sections_modular/05_fairness_and_bias_in_active_learning.tex
\textbf{Fairness and Bias in Active Learning:} recent research indicates that careless collection of data will exacerbate prevalent biases \cite{li2022more}. Label bias in data collection will be augmented by bias despite fairness constraints \cite{li2022more}. Uncertainty-based heuristics such as Bayesian Active Learning by Disagreement (BALD) enhance predictive parity and accuracy over i.i.d. sampling \cite{2104.06879}. New weights can correct and formalize statistical bias in AL \cite{2101.11665}.

Fair learning frameworks like Falcon employ multi-armed bandits for selecting informative examples and obtaining state-of-the-art performance \cite{10.14778/3641204.3641207}. Bayesian adaptive experimental design has the disadvantage of model misspecification bias, but the inclusion of noise in the model alleviates the problem \cite{2205.13698}. Novel weights correct statistical bias \cite{2101.11665}. Bayesian adaptive experimental design is evaluated for its robustness to bias, and experiments test the effectiveness of fair AL approaches \cite{2205.13698}. All works emphasize controlling label bias, statistical bias, and algorithmic bias in AL.

While there has been progress, challenges persist. It is challenging to deal with fairness and bias in AL and must account for several sources of bias \cite{li2022more}. Misspecification and noise impede the robustness of Bayesian active learning design \cite{2205.13698}. There is a trade-off between informativeness and postpone rate according to Falcon \cite{10.14778/3641204.3641207}. Scalability and efficacy are also crucial for fair AL methods \cite{2302.05711, 2312.08559}. Overall, the development of fair active learning approaches has significant implications for a wide range of applications, e.g., natural language processing \cite{2302.05711} and educational AI \cite{2407.18745}.

%% file: diagrams/ALBench.tex
\begin{figure*}[ht!]
\begin{center}
\includegraphics[width=0.9\linewidth]{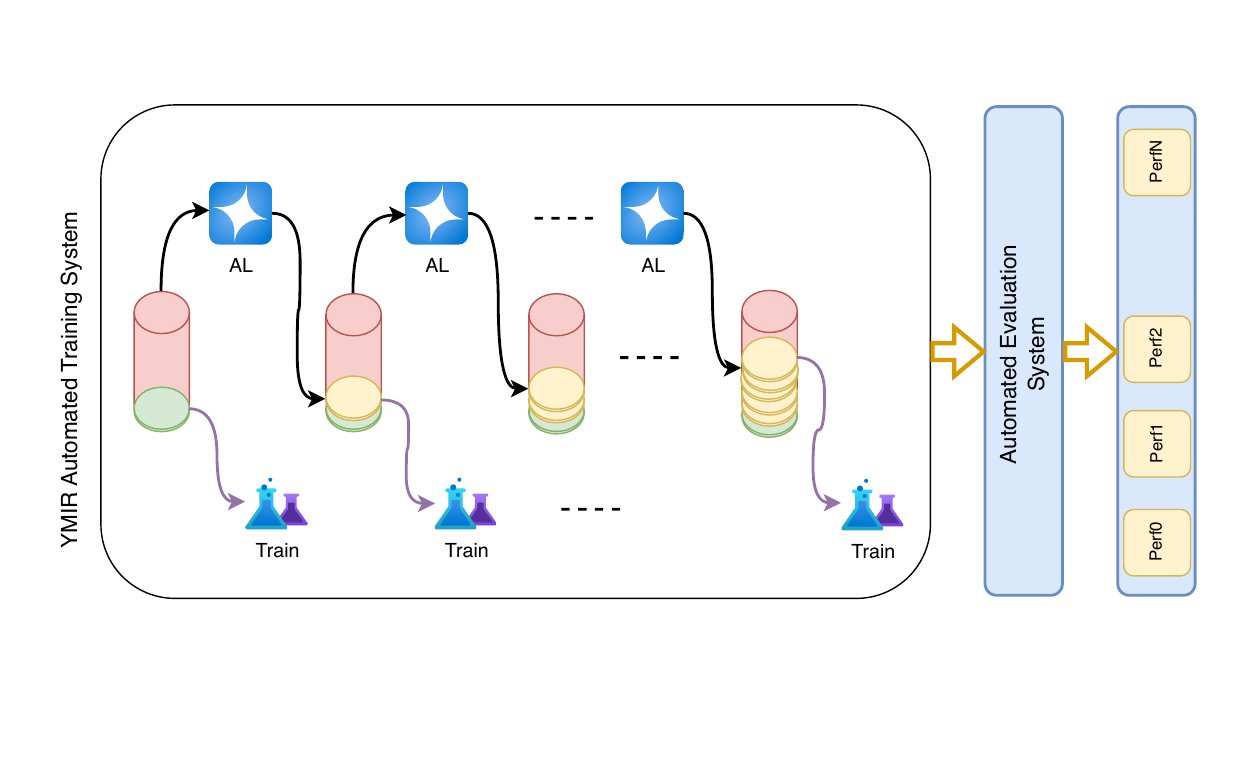}
\caption{Illustration of active learning procedure implemented by ALBench \cite{2207.13339} based on YMIR \cite{huang2021ymir}. Green indicates labeled data, yellow indicates mined data, and red indicates raw data.}
\end{center}
\end{figure*}

%% file: diagrams/AdaptiveSupervision_ObjectDetection.tex
\begin{figure*}[ht!]
\includegraphics[width=0.8\linewidth]{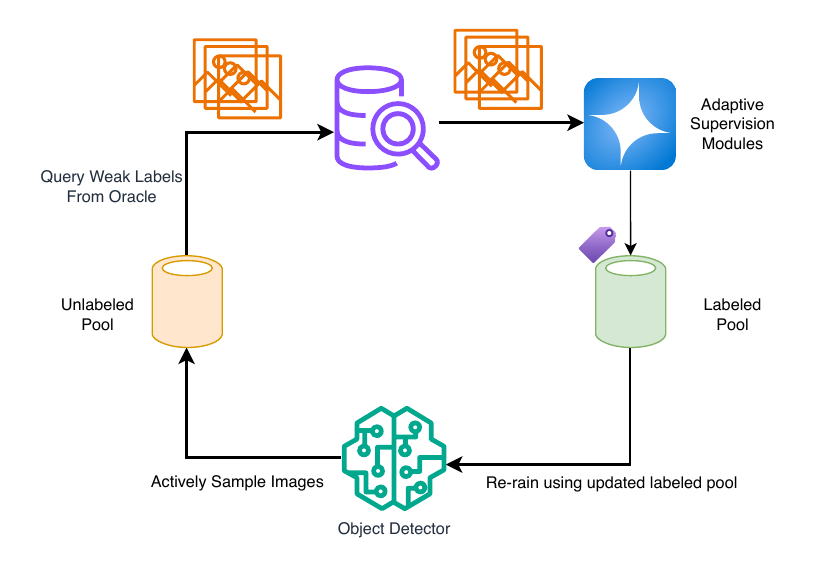}
\centering
\caption{Desai et al. [\citenum{1908.02454}] proposed framework integrating weak supervision into the active learning pipeline. It features an adaptive supervision module that dynamically escalates the level of supervision, enabling the transition to stronger supervision modes when necessary during model training.}
\end{figure*}

%% file: sections_modular/09_class_balance_and_imbalance_issues_in_active_learning.tex
\textbf{Class Balance and Imbalance Issues in Active Learning:} class imbalance hurts classifier performance in AL. It degrades the minority classes. Recent research pointed out that it is a potential factor in producing suboptimal classifiers and poor minority class performance \cite{bengar2022class, 2312.09196}. Researchers propose methods for handling class imbalance, including optimization frameworks that have a class-balancing feature \cite{bengar2022class}, illustrated below: 

\input{algorithms/class_balanced_AL}

\input{algorithms/greedy_class_balanced_AL}

Algorithm 1 (Class Balancing Active Learning) iteratively selects informative samples while ensuring class balance. It starts with a small labeled dataset and incrementally expands this set by querying new samples from an unlabeled pool. At each iteration, the method trains a CNN classifier and computes class-specific sampling thresholds $\Omega(c)$ defined as:
\input{algorithms/class_balanced_AL_omega_eq}

Minimizing the cost in Eq. \ref{eq:objective} encourages selecting a sufficient number of samples per class while choosing the most informative ones. 
\input{algorithms/class_balanced_AL_objective_eq}
The cost function consists of an affine term and an $\ell_1$ norm that are both convex, and subsequently, their linear combination is also convex. However, as the constraint is non-convex, the optimization problem becomes non-convex. The underlying problem is Binary Programming \cite{soudi1998optimized} that can be optimally solved by an off-the-shelf optimizer using LP relaxation \cite{yanover2006linear} and the branch-and-bound method. 

Algorithm 2 (Greedy Class Balancing Selection) provides an efficient, iterative approach to sample selection. At each step, the algorithm identifies the most informative sample by balancing two criteria: selecting samples far from existing labeled samples (based on a distance metric $D$), and maintaining class balance (guided by threshold $\Omega(c)$). Specifically, the algorithm greedily minimizes an objective that combines these two criteria. After each selection, the algorithm updates the distance matrix and continues until the budget per cycle is reached. In addition, others also include algorithms that spot class separation thresholds and pick the most uncertain instances \cite{2312.09196}, and graph-based techniques merge graph-based AL and deep learning \cite{zhang2022galaxy}.

Studies highlight the necessity of addressing various types of imbalance, such as proportion, variance, distance, neighborhood, and quality imbalance between/among classes \cite{2305.03900}. They enumerate types of imbalance: proportion, variance, distance, neighborhood, and quality. Real-world problems frequently encounter such imbalances \cite{2210.04675, zhao2021active}. Current AL techniques may not perform or may deteriorate on imbalanced data, highlighting the necessity for new techniques and approaches \cite{zhang2022galaxy, 2305.03900}.

Graph and deep learning methods are also under investigation as possible solutions for class imbalance in AL, with research showing their value in such applications as image classification \cite{bengar2022class} and NLP tasks \cite{2210.04675}. The method of choice will vary with task and data \cite{choi2021vab, gilhuber2023overcome}. Researchers are also investigating a variety of technical solutions, such as optimization frameworks \cite{bengar2022class}, new algorithms based on separation thresholds \cite{2312.09196}, and meta-learning for learning hyperparameters pertaining to imbalance \cite{2305.03900}.

Certain works propose new algorithms for treating imbalanced AL \cite{2312.09196, zhang2022galaxy}. Others provide theoretical suggestions and promote a rethink of class imbalance \cite{2305.03900}. There are different opinions regarding the necessity of class balance, with some believing that it’s crucial \cite{zhang2022galaxy}, and others proposing that there are ways of handling class imbalance by other means \cite{2312.09196}. 

Finally, effective solutions for managing class imbalance in AL will depend on further insight into the intricate relationships between learning strategies, class balance, and characteristics of the data. Effective methods necessitate further insight into how learning strategies, class balance, and data characteristics interplay. Uncovering these relationships will provide stronger systems for real-world tasks \cite{1407.1124, aaboud2018search}. As technology advances, we anticipate improved methods for coping with class imbalance. Better methods will enhance performance on applications.

%% file: algorithms/class_balanced_AL.tex
\begin{algorithm}[t]
\caption{Class Balancing AL [\citenum{bengar2022class}]}
\begin{algorithmic}
\renewcommand{\algorithmicrequire}{\textbf{Input:}}
\Require Unlabeled Pool $\mathcal{D_U}$, Total Budget $B$, Budget Per Cycle $b$  
\renewcommand{\algorithmicrequire}{\textbf{Initialize:}}
\Require Initial labeled pool $|\mathcal{D_L}|=b_{0} , c=1$   
\While {$|\mathcal{D_L}| < B$}
    \State Train CNN classifier $\Theta$ on $\mathcal{D_L}$
    \State Use $\Theta$ to compute probabilities for $x \in \mathcal{D_U}$ 
    \State Compute $\Omega(c) $ from Eq. \ref{eq:omega}
    \State Solve Eq. \ref{eq:objective} or Algorithm 2 for greedy, to obtain $z$
    \State Query $z$ to $\mathcal{ORACLE}$ 
    \State $\mathcal{D_L} \gets \mathcal{D_L} \cup z, \quad  \mathcal{D_U} \gets \mathcal{D_U} \setminus z$
    \State $c \gets c+1$
\EndWhile
\State\Return {$\mathcal{D_L}, \Theta$}
\end{algorithmic}
\end{algorithm}

%% file: algorithms/greedy_class_balanced_AL.tex
\begin{algorithm}[t]\label{algGreedy}
\caption{Greedy Class Balancing Selection [\citenum{bengar2022class}]}
\begin{algorithmic}
\renewcommand{\algorithmicrequire}{\textbf{Input:}}
\Require Softmax output $P_{N \times C}$, Distance Matrix $D_{N\times L}$, Balancing threshold $\Omega_{C\times 1}$, Regularizer $\lambda$, Budget Per Cycle $b$ 
\renewcommand{\algorithmicrequire}{\textbf{Initialize:}}
\Require $z^{(0)}=\mathbf{0}_{N\times1},\quad Q=P$  
\For{$i=0:b-1$}
    \vspace{2mm}
    \State $d^{(i)}_{N\times1}\gets	min(D,axis=1) \quad \triangleright$ for each unlabeled sample find the nearest labeled sample
    \vspace{2mm}
    \State $\psi\gets argmin(-d^{(i)}_{(N-i)\times1} +  \lambda \|\Omega(c)-Q^{T}_{C\times(N-i)}-P^{T}_{C\times N}z^{(i)}\mathds{1}_{1\times(N-i)})\|^{T}_1$
    \vspace{2mm}
    \State $z^{(i+1)}(\psi) \gets 1 \qquad \triangleright$ select the sample
    \State $Q\gets	P(z^{(i)}=0,:) \quad \triangleright$ keep the remaining unlabeled samples in $Q$
    \vspace{1mm}
    \State $D \gets D_{(N-i)\times (L+i)} \quad \triangleright$ update $D$ by removing a row and adding a column correspond to newly selected sample 
\EndFor
\State \Return $z^{(b)}$
\end{algorithmic}
\end{algorithm}

%% file: algorithms/class_balanced_AL_omega_eq.tex
\begin{equation} \label{eq:omega}
\Omega(c) =[\omega_{1},\omega_{2}, ...,\omega_{C}],
\end{equation}
\begin{equation} \label{eq:thresh}
\omega_{i}=max(\frac{cb+b_{0}}{C}-n_{i},0),
\end{equation}
where $\lambda$ is a parameter that regularizes the contribution of the balancing term in the objective.

%% file: algorithms/class_balanced_AL_objective_eq.tex
\begin{equation} \label{eq:objective}
\begin{aligned}
\min_{z}\quad
\boldsymbol{z}^{T}(P\odot \log{(P)}) \mathds{1}_{C\times 1}+\lambda\| \Omega(c)-P^T \boldsymbol{z} \|_{1},\\
\textrm{s.t.} \quad \boldsymbol{z}^{T} \mathds{1}_{N\times1}=b, \quad z_i \in \{0,1\}, \quad \forall i =1,2,...,N,\\
\end{aligned}
\end{equation}
where $\lambda$ is a parameter that regularizes the contribution of the balancing term in the objective. 

%% file: sections_modular/10_evaluation_metrics_and_benchmarks_for_active_learning.tex
\textbf{Evaluation Metrics and Benchmarks for Active Learning:} there have been calls from recent research for sound and reproducible evaluation, with appropriate re-benchmarking of available benchmarks in order to yield more sound results \cite{2306.08954}. New benchmarks like ALdataset \cite{2010.08161}, OpenAL \cite{2304.05246}, and CDALBench \cite{2408.00426} facilitate end-to-end comparisons.

A central theme is the recommendation of practical metrics for restoring trust in AL among such studies \cite{abraham2020rebuilding}. Metrics like interpretability metrics and statistical analysis techniques provide greater insight into AL performance. Cross-domain validation and large-scale reproducibility guarantee that methods generalize from one dataset \cite{2408.00426}. Furthermore, interpretability and transparency are essential. Studies demand metrics that explain model behavior \cite{2304.05246, abraham2020rebuilding}. Open experimentation environments and shared code enhance reproducibility \cite{2306.08954, 2010.08161}.

Perspectives on metrics and settings differ. Some researchers aim to design new benchmarks and evaluation metrics \cite{2010.08161}, e.g., OpenAL, proposed by Jonas et al.\ \cite{2304.05246}. Others advocate using existing metrics with new analysis techniques \cite{abraham2020rebuilding}. Experiment settings also vary between studies, from differences in the choice of datasets, oracle implementation, and repetition count \cite{2408.00426, 2306.08954}. Methodologies vary from benchmark design to statistical analysis. The design of new benchmarks like ALdataset and OpenAL necessitates attention to experimental settings and evaluation metrics \cite{2010.08161, 2304.05246}. Open-sourced implementations as well as efficient oracle implementations are also important in enabling the comparison of AL approaches \cite{2306.08954, 2408.00426}.

In spite of the advancements made in the construction of evaluation metrics and benchmarks for AL, there are still limitations and challenges. A deficiency in the standardization of benchmarks and evaluation metrics prevents comparison of AL approaches, and variability in the experimental conditions can give rise to conflicting results \cite{2306.08954, 2010.08161, 1108.0453, 1504.02406}. The necessity of performing large-scale evaluation with repeated runs highlights the importance of making thorough and trustworthy assessments of AL approaches \cite{2408.00426}.

In summary, working towards evaluation metrics and benchmarks for AL remains an important research focus, with themes such as the necessity for trustworthy benchmarks, interpretability, cross-domain evaluation, and reproducibility. Metrics and setups for experiments vary among researchers. The technical approaches from these works provide a strong base for subsequent research. Overcoming the limitations and challenges of these works will be vital in furthering our knowledge of AL strategies and their applications in other domains \cite{grimm2015self, iyyer2017amazing}.

%% file: sections_modular/03_applications_of_active_learning_in_computer_vision.tex
\textbf{Applications of Active Learning in Computer Vision:} AL is becoming more popular in computer vision because it helps reduce the cost of data labeling \cite{takezoe2023deep}. Many recent studies focus on lowering annotation costs while keeping training efficient \cite{tuia2012learning, 1908.02454}. For example, learning how confident a user is when labeling can help avoid asking for labels that are not necessary, which improves the overall process \cite{tuia2012learning}. Some methods first use weak labels and only request strong labels later. This helps save effort in tasks like object detection \cite{1908.02454}.

Active learning has many applications in computer vision, e.g. image classification, object detection, and semantic segmentation \cite{2012.04225, 2207.13339}. New methods aim to make active learning more effective and faster, including uncertainty sampling, query-by-committee, and reinforcement learning \cite{2108.05595, kyrkou2020imitation}. Several papers also discuss how to evaluate active learning better. They suggest new evaluation metrics and testing protocols for realistic comparisons \cite{10.1007/978-3-642-41398-8_16, 1912.05361}.

Take Greedy Active Learning (GAL) for example. Bar et al. \cite{2412.02310} propose a \emph{greedy-based} active learning framework that operates in two stages during each AL cycle. Denote the classifier at iteration $t$ as $\mathcal{C}_{t}$. In the initial stage, a candidate subset $X_c \subseteq X_u$ of size $K := |X_c|$ is selected from the unlabeled data pool. This subset may either encompass the entire unlabeled set or be filtered based on the top-K relevance scores. Due to inherent class imbalance, $X_c$ typically contains many irrelevant samples. In the following stage, a batch $X_b \subset X_c$ is chosen using an AL strategy. A human oracle then labels the instances in $X_b$, updating the labeled set $(X_l, \mathcal{Y}_l)$ with their features and labels. A new classifier $\mathcal{C}_{t+1}$ is trained on this augmented dataset for the next iteration.

Researchers are questioning traditional active learning methods. Some propose new frameworks that combine semi-supervised learning and data augmentation \cite{1912.05361}. Others explore combining active learning with transfer learning and self-supervised learning \cite{2301.01531, qu2025recent}. Deep learning models like CNNs and visual transformers are also widely used in active learning for computer vision tasks \cite{jeon2017active, 2106.03801}.

Despite recent progress, some challenges still exist. It is hard to scale active learning to very large datasets or complex problems \cite{parvaneh2022active, 2412.02310}. Also, poor-quality data or labels can hurt performance. Researchers are working on more efficient algorithms to solve these issues \cite{2202.01402}. High-quality datasets and better annotation protocols are also important for success \cite{bosser2021model}.

In conclusion, AL plays an important role in computer vision. It helps reduce labeling costs and improve learning efficiency. Recent work explores new evaluation methods and combines active learning with other learning techniques. Researchers continue to improve active learning approaches \cite{bang2024active, 2406.03845} to mitigate the challenges. The field is growing, and active learning will likely become even more important in future computer vision systems.

%% file: algorithms/GAL/GAL.tex
\newcommand{\etal}{\textit{et al}.~}
\newcommand{\ie}{\textit{i}.\textit{e}.~}
\newcommand{\eg}{\textit{e}.\textit{g}.~}
\newcommand{\X}{\mathcal{X}}
\newcommand{\F}{\mathcal{F}}
\newcommand{\A}{\mathcal{A}}
\newcommand{\CommentTriangle}[1]{\hfill$\triangleright$ #1}
\algrenewcommand\alglinenumber[1]{}  

\begin{algorithm}
\caption{{Greedy Active Learning (GAL) Algorithm \cite{2412.02310}}}
\label{alg:greedy_gen}

\begin{algorithmic}
\Function{GAL}{$\mathcal{X}_c, \mathcal{X}_l, \mathcal{Y}_l, B$}
    \State $\mathcal{X}_b \gets \{\}$
    \For{$i \gets 1$ to $B$}
        \State $x^*, \hat{l}^* \gets \Call{Next}{\mathcal{X}_c, \mathcal{X}_l, \mathcal{Y}_l}$ \Comment{Find the point that maximizes impact $\mathcal{S}$}
        \State $\mathcal{X}_l \gets \mathcal{X}_l \cup \{x^*\}$
        \State $\mathcal{Y}_l \gets \mathcal{Y}_l \cup \{\hat{l}^*\}$
        \State $\mathcal{X}_c \gets \mathcal{X}_c \setminus x^*$
        \State $\mathcal{X}_b \gets \mathcal{X}_b \cup \{x^*\}$
    \EndFor
    \State \Return $\mathcal{X}_b$
\EndFunction

\vspace{1em}

\Function{Next}{$\mathcal{X}_c, \mathcal{X}_l, \mathcal{Y}_l$}
    \For{$i \gets 1$ to $|\mathcal{X}_c|$}
        \State $x_i \gets \mathcal{X}_c[i]$
        \State $\mathcal{S}_i, \hat{l}_i \gets \psi(x_i, \mathcal{X}_l, \mathcal{Y}_l)$ \Comment{Acquisition function (see Alg.~\ref{alg:sel})}
    \EndFor
    \State $i^* \gets \arg\max_i \mathcal{S}_i$
    \State \Return $x_{i^*}, \hat{l}_{i^*}$ \Comment{Return optimal point and pseudo label}
\EndFunction
\end{algorithmic}
\end{algorithm}

%% file: algorithms/GAL/Aquisition_function.tex
\begin{algorithm}
\caption{{Acquisition Functions \cite{2412.02310}}}
\label{alg:sel}
\begin{algorithmic}
\Function{$\psi_{svm}$}{$x_i$, $\mathcal{X}_l$, $\mathcal{Y}_l$}\Comment{SVM}

        \State {$\theta^+ \gets \text{Classifier}(\mathcal{X}_l \cup x_i$, $\mathcal{Y}_l \cup +1)$}
        
        \State {$\theta^- \gets \text{Classifier}(\mathcal{X}_l \cup x_i$, $\mathcal{Y}_l \cup -1)$}
        
        \State {$\hat{l}_i \gets \argmin_{l_i\in\{-1,+1\}}\mathcal{F}_{svm}(x_i,l_i,\theta^{l_i})$}\Comment{Eqs. \ref{eq:svm}, \ref{eq:fsvm}}
        
        \State {$\mathcal{S}_i \gets \mathcal{F}_{svm}(x_i,\hat{l}_i,\theta^{\hat{l}_i})$} 
     \State \Return $\mathcal{S}_i, \hat{l}_{i}$
     \EndFunction

 \State
 \Function{$\psi_{gp}$}{$x_i$, $\mathcal{X}_l$, $\mathcal{Y}_l$}\Comment{Gaussian Process}
         \State {$\mathcal{S}_i \gets \mathcal{F}_{gp}
         (x_i,\X_l)$} by~\eqref{eq:gp}
         \State \Return $\mathcal{S}_i, Null$
 \EndFunction 

\end{algorithmic}
\end{algorithm} 

%% file: sections_modular/04_active_learning_for_natural_language_processing_tasks.tex
\textbf{Active Learning for Natural Language Processing Tasks:} many papers show AL can help improve data efficiency and model accuracy \cite{zhang2022survey, 2104.08320, 1810.03450}. One important idea is adapting pre-trained language models to the specific task. Training more on unlabeled data can improve results and reduce the need for labels \cite{2104.08320}. For example, ensemble methods like Majority-CRF \cite{1810.03450} are used in natural language understanding. They reduce error rates by 6.6\%–9\% compared to random sampling. Another method, active query k-means, also helps increase accuracy and cut training costs in text classification \cite{jiang2021application}.

Model adaptation is a crucial topic in NLP active learning. It shows the importance of learning strategies for pre-trained models \cite{zhang2022survey, 2104.08320}. Ensemble methods are popular too, as they help pick the most useful samples \cite{1810.03450}. Semi-supervised learning is often used with active learning to reduce labeling needs \cite{kohl2024scoping, jiang2021application}. Beatty et al. show that choosing when to stop active learning using labeled vs. unlabeled data is important, especially for text classification \cite{beatty2019use}. Good evaluation tools, e.g., datasets and metrics, are also important. Most papers use F1-score to measure performance \cite{kohl2024scoping}. The corresponding equations in pseudocodes are:

\input{algorithms/GAL/eqs}

There are different opinions on which query strategy is best. Some focus on exploitation, some on exploration, and others mix both \cite{kohl2024scoping}. Some researchers prefer continuous training on unlabeled data to adapt Language Models (LMs). Others say fine-tuning works better: Margatina et al. \cite{2104.08320} propose \textit{AL with Pretrained LMs}: Starting with a pretrained LM $\mathcal{P}(x;W_0)$, the model undergoes task-adaptive pretraining on the unlabeled pool $D_\text{pool}$ to obtain $\mathcal{P}_{\text{TAPT}{}}(x;W_0')$. In each AL round, a classifier $\mathcal{M}_i$ is initialized from $\mathcal{P}_{\text{TAPT}{}}$ and fine-tuned on the labeled set $D_\text{lab}$. A batch is selected using the acquisition function $a$, and the process repeats.

Active query k-means offers another option beyond traditional methods \cite{jiang2021application}. Many sampling strategies are proposed, such as uncertainty sampling, query-by-committee, and Penalized Min-Max-selection \cite{zhang2022survey, 1810.03450, jiang2021application}.

AL in NLP still has open problems. One issue is the lack of strong comparisons between methods. This makes it hard to know which strategy is best \cite{kohl2024scoping}. Also, labeling NLP data takes time and money, especially for large tasks \cite{zhang2022survey, 1810.03450}. Adapting language models to new tasks is also not easy and needs more research \cite{2104.08320}. Recent studies look at active learning for sequence labeling \cite{humeniuk2023ambiegen} and machine translation \cite{virtanen2022product}. These show that active learning can help in many NLP tasks.

To sum up, active learning shows strong potential in NLP. It helps save labels and improve results. But more research is needed to solve current challenges. New methods like active query k-means and better model adaptation can lead to better results \cite{zhang2022survey, 2104.08320, 1810.03450}. In the future, active learning could help build more accurate and cost-efficient NLP models \cite{kohl2024scoping, jiang2021application}.

%% file: algorithms/GAL/eqs.tex
\begin{equation}
\mathcal{F}_{svm} := \|W(x_i,l_i)-W_0\|^2_2,
\label{eq:svm}
\end{equation}

\begin{equation}
\mathcal{S}_i=\min_{l_i\in\{-1,+1\}} \mathcal{F}_{svm}(x_i,l_i,\theta^{l_i}),
\label{eq:fsvm}
\end{equation}

\begin{equation}
\mathcal{{F}}_{gp}(x_i) := -\Big(\sum_{x \in X_c}\sigma_{A\cup x_i}^2(x)+\alpha\max_{x \in X_c}\sigma_{A\cup x_i}^2(x)\Big).
\label{eq:gp}
\end{equation}

%% file: algorithms/Margatina_2022/Margatina_2022.tex
\newcommand{\Dpool}{\mathcal{D}_{\textbf{pool}}}
\newcommand{\Dlab}{\mathcal{D}_{\textbf{lab}}}
\newcommand{\Dval}{\mathcal{D}_{\textbf{val}}}
\newcommand{\Dtest}{\mathcal{D}_{\textbf{test}}}

\newcommand{\ft}{\textsc{ft+}}
\newcommand{\tapt}{\textsc{tapt}}

\begin{algorithm}[!t]
\caption{Active Learning with Pretrained LMs \cite{2104.08320}}
\label{algo:balm}
\begin{algorithmic}[1]

\State \textbf{Input:} Unlabeled pool $\Dpool$; pretrained LM $\mathcal{P}(x;W_0)$; acquisition size $k$; AL iterations $T$; acquisition function $a$
\State \textbf{Output:} Labeled dataset $\Dlab$

\State $\Dlab \gets \emptyset$
\State $\mathcal{P}_{\text{TAPT}}(x;W_0') \gets$ Train $\mathcal{P}(x;W_0)$ on $\Dpool$
\State $\mathcal{Q}_0 \gets$ \textsc{Random}$(\cdot), \; |\mathcal{Q}_0| = k$
\State $\Dlab \gets \Dlab \cup \mathcal{Q}_0$
\State $\Dpool \gets \Dpool \setminus \mathcal{Q}_0$

\For{$i \gets 1$ to $T$}
    \State $\mathcal{M}_i(x;[W_0',W_c]) \gets$ Initialize from $\mathcal{P}_{\text{TAPT}}(x;W_0')$
    \State $\mathcal{M}_i(x;W_i) \gets$ Train model on $\Dlab$
    \State $\mathcal{Q}_i \gets a(\mathcal{M}_i, \Dpool, k)$
    \State $\Dlab \gets \Dlab \cup \mathcal{Q}_i$
    \State $\Dpool \gets \Dpool \setminus \mathcal{Q}_i$
\EndFor

\State \Return $\Dlab$
\end{algorithmic}
\end{algorithm}

%% file: diagrams/AADA.tex
\begin{figure}
\includegraphics[scale=0.9]{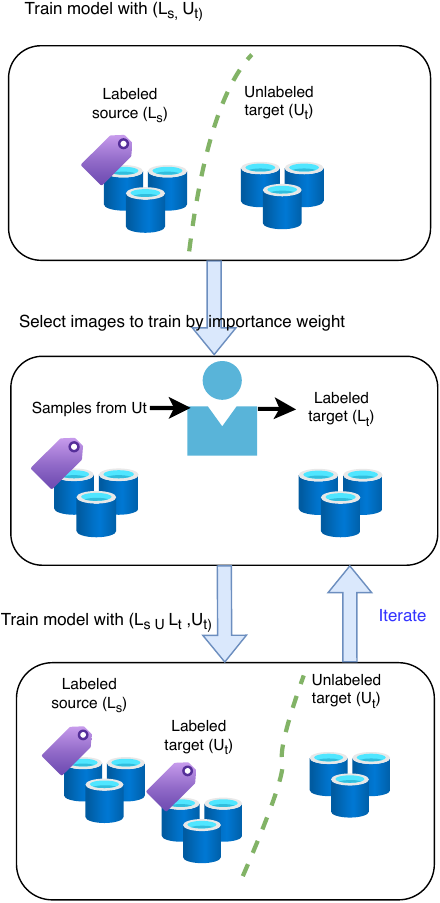}
\centering
\caption{Illustration of \textbf{AADA} \cite{su2020active} algorithm. The algorithm operates within the unsupervised domain adaptation framework. Initially, the authors train a model using the labeled source data ($L_s$) and the unlabeled target data ($U_t$), leveraging a domain adversarial loss. In each subsequent iteration, the authors select informative samples from the target domain based on importance weighting and acquire their labels. The model is then re-trained using the combined labeled data ($L_s \cup L_t$) along with the remaining unlabeled target data ($U_t$).}
\end{figure}

%% file: diagrams/Empowering.tex
\begin{figure*}[ht!]
\includegraphics[width=0.95\linewidth]{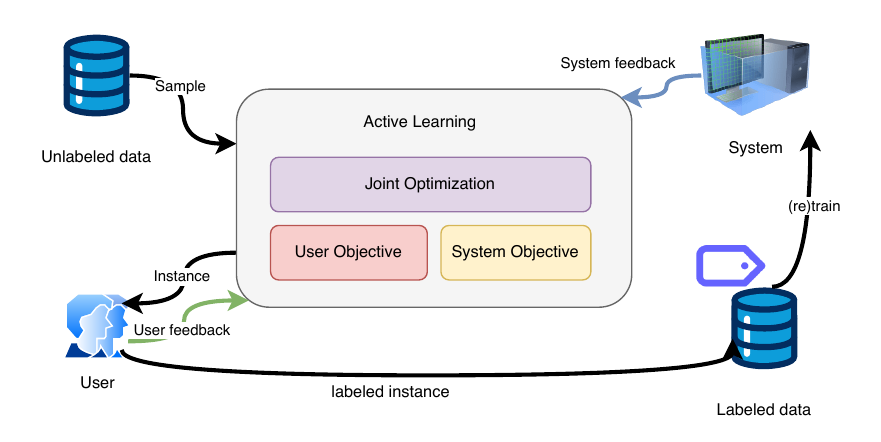}
\caption{An overview of the interactive framework by Lee et al. \cite{2005.04470}. Unlike prior approaches that focus solely on optimizing the system objective (\textcolor{blue}{blue modules}), their method also incorporates the user objective (\textcolor{green!60!black}{green line}), enabling a joint optimization of these potentially conflicting goals (\textcolor{yellow!60!black}{yellow block}).}
\end{figure*}

%% file: sections_modular/06_domain_adaptation_and_transfer_learning_in_active_learning.tex
\textbf{Domain Adaptation and Transfer Learning in Active Learning:} domain adaptation and transfer learning enable models to perform competently in different domains and tasks. Several recent works have attempted to resolve challenges such as domain shift and knowledge transfer. Active Adversarial Domain Adaptation (AADA) \cite{su2020active} utilizes adversarial domain alignment and importance sampling within a unified framework. It chooses samples based on their uncertainty and diversity, indicating that these two aspects play a crucial role in active learning.

The domain adaptation survey for visual applications \cite{1702.05374} provides a decent summary of techniques in image classification and object detection. It covers both shallow and deep learning methods. There exists a survey on the theory of domain adaptation \cite{2004.11829} that deals with learning bounds and theoretical guarantees. It indicates that there should be more analysis and theory for the field.

A number of new approaches have also been proposed. Divide and Adapt (DiaNA) \cite{huang2023divide} constructs four target groups of different transferability. It adapts each of these groups using a particular approach, and the performance improves. Adaptive Feature Norm \cite{xu2019larger} demonstrates that features with higher norms transfer optimally. This approach outperforms other state-of-the-art methods for unsupervised domain adaptation.

Most of the papers concur that domain adaptation plays a crucial role in active learning. They also mention how one should select quality samples for labeling. Transfer learning helps transfer knowledge across tasks and domains. Uncertainty and diversity are shared techniques for sampling. We observe the same in AADA \cite{su2020active}, DiaNA \cite{huang2023divide}, and other such tasks like Discriminative Active Learning \cite{zhou2021discriminative} and Domain Adversarial Reinforcement Learning \cite{chen2020domain}.

Varying technical methods are utilized. Adversarial alignment \cite{su2020active} is applied by AADA. Importance sampling and Gaussian Mixture Models are utilized by DiaNA \cite{huang2023divide}. Adaptive Feature Norm adjusts the norms of features so that domains become similar \cite{xu2019larger}. Learning to Adapt \cite{dakic2017macroscopic} and Domain-Invariant Feature Learning \cite{1310.5195} are other methods that reduce domain shift. Despite progress, there remain challenges. One of them is negative transfer, in which performance degrades following adaptation \cite{1702.05374, 2004.11829}. 

Challenges remain in domain adaptation: Certain methods are slow or do not generalize effectively to big data \cite{xu2019larger, huang2023divide}. The frequent lack of theoretical analysis constrains our knowledge \cite{2004.11829}. To treat overfitting or underfitting, we require regularization and thoughtful model selection \cite{su2020active, huang2023divide}.

In conclusion, transfer learning and domain adaptation are central to enabling active learning in different tasks. Several effective techniques have long been proposed for addressing domain shift and knowledge transfer. There remain some problems that still require work. Fresh ideas, such as in Towards Domain Adaptation \cite{baldassi2016unreasonable} and the Survey on Transfer Learning \cite{tan2018survey}, will keep advancing the field. All these advancements are crucial for making active learning more effective and useful.

%% file: diagrams/DEAL.tex
\begin{figure*}[ht!]
\includegraphics[width=0.78\linewidth]{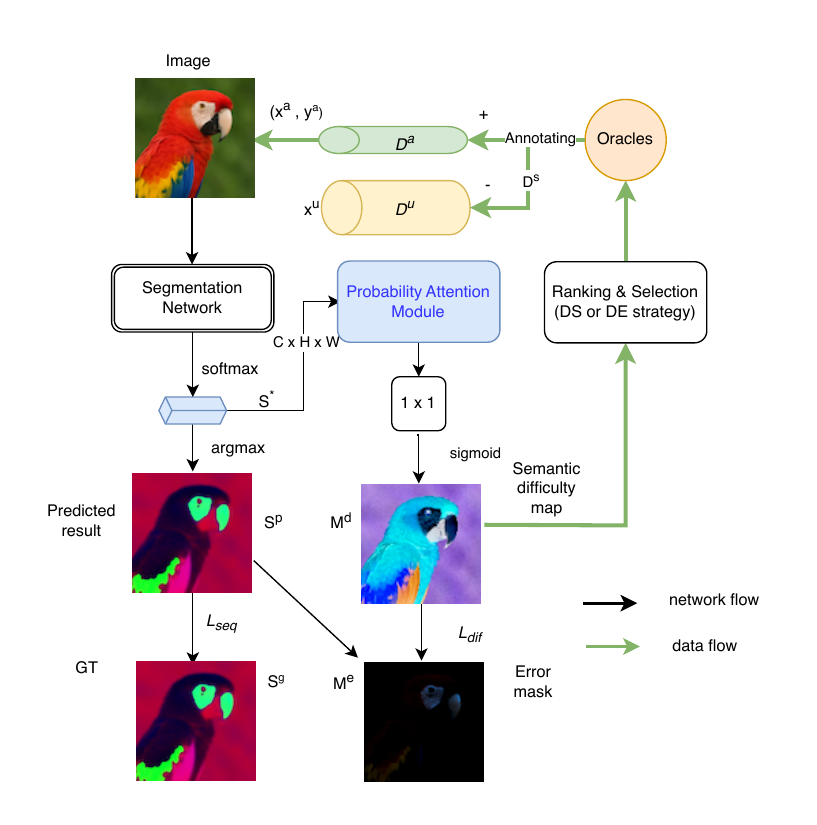}
\centering
\caption{Illustration of the DEAL \cite{xie2020deal}: difficulty-aware active learning framework for semantic segmentation. The architecture consists of two main branches: the first follows a standard semantic segmentation network, while the second incorporates a probability attention module followed by a \(1 \times 1\) convolution. Here, \(D^a\) and \(D^u\) represent the labeled and unlabeled datasets, respectively, and \(D^s\) denotes a selected subset from \(D^u\). Probability maps before and after the attention mechanism are denoted by \(P\) and \(Q\), respectively. The framework optimizes two loss functions, \(\mathcal{L}_{\text{seg}}\) and \(\mathcal{L}_{\text{dif}}\) which also introduced in this work \cite{xie2020deal}.}
\end{figure*}

%% file: sections_modular/11_realworld_applications_and_case_studies_of_active_learning.tex
\textbf{Real-World Applications and Case Studies of Active Learning:} in recent years, many papers have studied how active learning works in real-world applications. A simulation study in \cite{10.1007/978-3-642-41398-8_16} tests different active learning methods in various situations. It shows that we need clear and accurate metrics to measure performance. Another paper \cite{abraham2020rebuilding} introduces actionable metrics. These help people in industry better understand and evaluate active learning, making it easier to trust and apply. In education, \cite{2005.04470} shows how active learning can improve student engagement. These examples show that active learning is important for building strong AI models in different areas.

Balancing system performance with user needs is also an important topic. For example, \cite{2005.04470} proposes a method that considers both model quality and user demands. It works well in an educational setting. This kind of user-focused approach helps make active learning more reliable in real-world use. But other papers focus more on improving system performance, showing a trade-off between system goals and user experience \cite{bosser2021model}. Simulation experiments like \cite{1408.1319} help test model performance. Query strategies from \cite{zhang2022survey} are key to selecting useful samples.

However, there exist many challenges. One big challenge is exploration—active learning sometimes struggles to search the full data space \cite{1206.5274}. Also, picking the best samples to label is not easy \cite{2011.03733}. Some papers discuss how to handle AutoML systems \cite{bosser2021model}, and how to bring research ideas into practical industry use \cite{1907.00038}.

Active learning can be used in many industries. It also connects with other areas like reinforcement learning \cite{2201.03947} and human-like learning \cite{2011.03733}. As the research grows, we must solve open problems to make active learning useful. We need better metrics, stronger technical solutions, and a good balance between model performance and user needs. Future progress depends on solving these challenges \cite{2306.08001, 2403.14800}.

%% file: sections_modular/14_active_learning_for_specialized_tasks.tex
\textbf{Active Learning for Specialized Tasks:} AL is used more often in specialized tasks like semantic segmentation, object detection, and recognition nowadays. The goal is to get good results using only a small amount of labeled data. Many recent studies say that we need better and more realistic ways to evaluate active learning methods \cite{1912.05361}, \cite{2207.13339}. For example, the research shows that deep active learning may not always perform better than random sampling, and results can change a lot depending on how the model is trained \cite{1912.05361}. Another work, ALBench, builds a benchmark to help researchers compare methods fairly and reproduce results in object detection \cite{2207.13339}.

In semantic segmentation, active learning tries to focus on areas that are harder to segment. This helps improve results in those difficult regions \cite{xie2020deal}. The DEAL method \cite{xie2020deal} learns how to estimate which areas are hard to segment using pixel-wise attention scores. This shows the importance of designing active learning methods for the specific task. Other papers combine active learning with semi-supervised learning to gain performance on segmentation tasks \cite{2210.08403}, \cite{2203.10730}. But the performance depends on the task and the dataset.

In object detection, active learning is used to lower the cost of labeling and improve model accuracy \cite{goupilleau2021active}, \cite{wang2023alwod}. The ALWOD \cite{wang2023alwod} shows a method for active learning in weakly supervised object detection. It proves that active learning can help in this area too. Another study looks at how well low-cost methods work for object detection \cite{probst2022evaluating}. It points out the need for more efficient strategies.

Several papers also talk about real-world problems. They say active learning must adapt to changes in data and task conditions \cite{1907.00038}, \cite{mittal2023best}. Discwise AL for LiDAR semantic segmentation \cite{unal2023discwise} shows how active learning can be used in more types of data. The Best Practices paper gives advice on how to choose the right active learning method \cite{mittal2023best}. It considers factors like the type of data, whether semi-supervised learning is used, and how many labels are available.

To summarize, active learning is now widely used in tasks like segmentation, detection, and recognition. It helps improve results when labeled data is limited. But we still need better benchmarks \cite{2207.13339}, \cite{1912.05361}, more realistic tests, and smarter methods that handle specific tasks and real-world problems. Combining active learning with semi-supervised learning is one promising direction. Still, more research is needed to fix current issues and build faster, more reliable active learning systems \cite{1912.05361}, \cite{2207.13339}.

%% file: sections_modular/16_conclusion.tex
This paper presents a detailed overview of the recent developments in active learning. It surveys key findings, common patterns, and technical approaches from a wide range of studies. A major highlight is that active learning often performs better than passive learning. Especially, human-like active learning methods show strong potential in creating more effective training systems. Researchers have examined several components of deep active learning, including model architecture, query strategies, and the use of pseudo-labels. These studies help to enhance the overall performance of active learning. These findings suggest that AL can significantly boost machine learning. Some challenges remain: the lack of standardized benchmarks, fairness concerns, and class imbalance. These need to be addressed to achieve full effectiveness. Community's development on practical and trustworthy methods can support the application of active learning in real-world applications, e.g. computer vision, NLP, etc. In the future, stronger connections between AL and fields like deep learning and NLP will be essential for further progress.

%% file: cite.bib
@misc{1108.0453,
  author = {Vladimir Nikulin},
  title = {On the Evaluation Criterions for the Active Learning Processes},
  year = {2011},
  primaryClass = {stat.ML},
  url = {https://arxiv.org/abs/1108.0453}
}

@inproceedings{1206.5274,
  title={On Discarding, Caching, and Recalling Samples in Active Learning},
  author={Kapoor, Ashish and Horvitz, Eric},
  booktitle={Proceedings of the Twenty-Third Conference on Uncertainty in Artificial Intelligence (UAI)},
  pages={200--207},
  year={2007},
  publisher={AUAI Press},
  url = {https://arxiv.org/abs/1206.5274}
}

@article{kruschke2008bayesian,
  title={Bayesian approaches to associative learning: From passive to active learning},
  author={Kruschke, John K},
  journal={Learning \& behavior},
  volume={36},
  number={3},
  pages={210--226},
  year={2008},
  publisher={Springer}
}

@article{puterman1990markov,
  title={Markov decision processes},
  author={Puterman, Martin L},
  journal={Handbooks in operations research and management science},
  volume={2},
  pages={331--434},
  year={1990},
  publisher={Elsevier}
}

@inproceedings{li2024adaptive,
  title={Adaptive uncertainty-based learning for text-based person retrieval},
  author={Li, Shenshen and He, Chen and Xu, Xing and Shen, Fumin and Yang, Yang and Shen, Heng Tao},
  booktitle={Proceedings of the AAAI Conference on Artificial Intelligence},
  volume={38},
  pages={3172--3180},
  year={2024}
}

@article{carvalho2024deep,
  title={Deep bayesian active learning for preference modeling in large language models},
  author={Carvalho Melo, Luckeciano and Tigas, Panagiotis and Abate, Alessandro and Gal, Yarin},
  journal={Advances in Neural Information Processing Systems},
  volume={37},
  pages={118052--118085},
  year={2024}
}

@inproceedings{hernandez2015probabilistic,
  title={Probabilistic backpropagation for scalable learning of bayesian neural networks},
  author={Hern{\'a}ndez-Lobato, Jos{\'e} Miguel and Adams, Ryan},
  booktitle={International conference on machine learning},
  pages={1861--1869},
  year={2015},
  organization={PMLR}
}

@article{he2021towards,
  title={Towards non-iid image classification: A dataset and baselines},
  author={He, Yue and Shen, Zheyan and Cui, Peng},
  journal={Pattern Recognition},
  volume={110},
  pages={107383},
  year={2021},
  publisher={Elsevier}
}

@misc{1310.5195,
  author = {Liu, Chien-Hao and Yau, Shing-Tung},
  title = {A mathematical theory of D-string world-sheet instantons, II: Moduli stack of $Z$-(semi)stable morphisms from Azumaya nodal curves with a fundamental module to a projective Calabi-Yau 3-fold},
  year = {2013},
  url = {https://arxiv.org/abs/1310.5195}
}

@inproceedings{1312.2936,
  title={Active Player Modelling},
  author={Togelius, Julian and Shaker, Noor and Yannakakis, Georgios N.},
  booktitle={Proceedings of the 9th International Conference on the Foundations of Digital Games (FDG)},
  year={2014},
  url = {https://arxiv.org/abs/1312.2936}
}

@misc{1407.1124,
  author = {Mikovic, Aleksandar and Vojinovic, Marko},
  title = {Cosmological Constant in a Quantum Gravity Theory for a Piecewise-linear Spacetime},
  year = {2014},
  url = {https://arxiv.org/abs/1407.1124}
}

@incollection{1408.1319,
  title={When Does Active Learning Work?},
  author={Evans, Lewis and Adams, Niall M. and Anagnostopoulos, Christoforos},
  booktitle={Advances in Intelligent Data Analysis XIII},
  pages={63--75},
  year={2014},
  publisher={Springer},
  url = {https://arxiv.org/abs/1408.1319}
}

@article{1504.02406,
  title={Deciding when to stop: Efficient stopping of active learning guided drug-target prediction},
  author={Temerinac-Ott, Maja and Naik, Armaghan W. and Murphy, Robert F.},
  journal={BMC Bioinformatics},
  volume={16},
  number={1},
  pages={289},
  year={2015},
  publisher={BioMed Central},
  url = {https://arxiv.org/abs/1504.02406}
}

@misc{1702.05374,
  author = {Csurka, Gabriela},
  title = {Domain Adaptation for Visual Applications: A Comprehensive Survey},
  year = {2017},
  url = {https://arxiv.org/abs/1702.05374}
}

@inproceedings{bosser2021model,
  title={Model-centric and data-centric aspects of active learning for deep neural networks},
  author={Boss{\'e}r, John Daniel and S{\"o}rstadius, Erik and Chehreghani, Morteza Haghir},
  booktitle={2021 IEEE International Conference on Big Data (Big Data)},
  pages={5053--5062},
  year={2021},
  organization={IEEE}
}

@misc{1711.01732,
  author = {Lin, Xiao and Parikh, Devi},
  title = {Active Learning for Visual Question Answering: An Empirical Study},
  year = {2017},
  url = {https://arxiv.org/abs/1711.01732}
}

@misc{1803.06586,
  author = {Tosh, Christopher and Dasgupta, Sanjoy},
  title = {Structural query-by-committee},
  year = {2018},
  url = {https://arxiv.org/abs/1803.06586}
}

@inproceedings{1808.05697,
  title={Deep Bayesian Active Learning for Natural Language Processing: Results of a Large-Scale Empirical Study},
  author={Siddhant, Aditya and Lipton, Zachary C.},
  booktitle={Proceedings of the 2018 Conference on Empirical Methods in Natural Language Processing (EMNLP)},
  pages={2904--2909},
  year={2018},
  url = {https://arxiv.org/abs/1808.05697}
}

@misc{1810.03450,
  author = {Peshterliev, Stanislav and Kearney, John and Jagannatha, Abhyuday and Kiss, Imre and Matsoukas, Spyros},
  title = {Active Learning for New Domains in Natural Language Understanding},
  year = {2018},
  url = {https://arxiv.org/abs/1810.03450}
}

@misc{1907.00038,
  author = {Kagy, Jean-François and Kayadelen, Tolga and Ma, Ji and Rostamizadeh, Afshin and Strnadova, Jana},
  title = {The Practical Challenges of Active Learning: Lessons Learned from Live Experimentation},
  year = {2019},
  url = {https://arxiv.org/abs/1907.00038}
}

@inproceedings{1908.02454,
  title={An Adaptive Supervision Framework for Active Learning in Object Detection},
  author={Desai, Sai Vikas and Chandra, Akshay L and Guo, Wei and Ninomiya, Seishi and Balasubramanian, Vineeth N},
  booktitle={British Machine Vision Conference (BMVC)},
  year={2019},
  url = {https://arxiv.org/abs/1908.02454}
}

@misc{1909.04928,
  author = {Jethava, Vinay},
  title = {On Weighted Uncertainty Sampling in Active Learning},
  year = {2019},
  url = {https://arxiv.org/abs/1909.04928}
}

@inproceedings{1912.05361,
  title={Parting with Illusions about Deep Active Learning},
  author={Mittal, Sudhanshu and Tatarchenko, Maxim and Çiçek, Ozgun and Brox, Thomas},
  booktitle={Proceedings of the IEEE/CVF Conference on Computer Vision and Pattern Recognition (CVPR)},
  pages={9433--9442},
  year={2019},
  doi={10.1109/CVPR.2019.00966},
  url = {https://arxiv.org/abs/1912.05361}
}

@misc{2004.11829,
  author = {Redko, Ievgen and Morvant, Emilie and Habrard, Amaury and Sebban, Marc and Bennani, Younès},
  title = {A Survey on Domain Adaptation Theory: Learning Bounds and Theoretical Guarantees},
  year = {2020},
  url = {https://arxiv.org/abs/2004.11829}
}

@inproceedings{2005.04470,
  title={Empowering Active Learning to Jointly Optimize System and User Demands},
  author={Ji-Ung Lee and Christian M. Meyer and Iryna Gurevych},
  booktitle={Proceedings of the 58th Annual Meeting of the Association for Computational Linguistics},
  pages={4334--4345},
  year={2020},
  publisher={Association for Computational Linguistics},
  doi={10.18653/v1/2020.acl-main.390},
  url = {https://arxiv.org/abs/2005.04470}
}

@inproceedings{2005.07402,
  title={Stopping Criterion for Active Learning Based on Deterministic Generalization Bounds},
  author={Hideaki Ishibashi and Hideitsu Hino},
  booktitle={Proceedings of the 23rd International Conference on Artificial Intelligence and Statistics},
  pages={3486--3496},
  year={2020},
  publisher={PMLR},
  url = {https://arxiv.org/abs/2005.07402}
}

@inproceedings{2010.08161,
  author = {Xueying Zhan and Antoni B. Chan},
  title = {ALdataset: A Benchmark for Pool-Based Active Learning},
  booktitle = {Proceedings of the 30th International Joint Conference on Artificial Intelligence (IJCAI)},
  pages = {4679--4686},
  year = {2021},
  url = {https://arxiv.org/abs/2010.08161}
}

@inproceedings{2011.03733,
  author = {Jaeseo Lim and Hwiyeol Jo and Byoung-Tak Zhang and Jooyong Park},
  title = {Human-Like Active Learning: Machines Simulating the Human Learning Process},
  booktitle = {Proceedings of the 34th Conference on Neural Information Processing Systems (NeurIPS)},
  year = {2020},
  url = {https://arxiv.org/abs/2011.03733}
}

@inproceedings{2012.04225,
  title={Active Learning: Problem Settings and Recent Developments},
  author={Hino, Hideitsu},
  booktitle={arXiv preprint arXiv:2012.04225},
  year={2020},
  url={https://arxiv.org/abs/2012.04225}
}

@inproceedings{2101.11665,
  title={On Statistical Bias In Active Learning: How and When To Fix It},
  author={Farquhar, Sebastian and Gal, Yarin and Rainforth, Tom},
  booktitle={International Conference on Learning Representations (ICLR)},
  year={2021},
  url={https://arxiv.org/abs/2101.11665}
}

@inproceedings{2104.06879,
  title={Can Active Learning Preemptively Mitigate Fairness Issues?},
  author={Branchaud-Charron, Frédéric and Atighehchian, Parmida and Rodríguez, Pau and Abuhamad, Grace and Lacoste, Alexandre},
  booktitle={arXiv preprint arXiv:2104.06879},
  year={2021},
  url={https://arxiv.org/abs/2104.06879}
}

@inproceedings{2104.08320,
  title={On the Importance of Effectively Adapting Pretrained Language Models for Active Learning},
  author={Margatina, Katerina and Barrault, Loïc and Aletras, Nikolaos},
  booktitle={Proceedings of the 60th Annual Meeting of the Association for Computational Linguistics (ACL)},
  pages={4388--4397},
  year={2022},
  publisher={Association for Computational Linguistics},
  url={https://arxiv.org/abs/2104.08320}
}

@inproceedings{2106.03801,
  title={Visual Transformer for Task-aware Active Learning},
  author={Caramalau, Razvan and Bhattarai, Binod and Kim, Tae-Kyun},
  booktitle={arXiv preprint arXiv:2106.03801},
  year={2021},
  url={https://arxiv.org/abs/2106.03801}
}

@article{2108.05595,
  title={Reinforcement Learning Approach to Active Learning for Image Classification},
  author={Werner, Thorben},
  journal={arXiv preprint arXiv:2108.05595},
  year={2021}
}

@inproceedings{2202.01402,
  title={GALAXY: Graph-based Active Learning at the Extreme},
  author={Zhang, Jifan and Katz-Samuels, Julian and Nowak, Robert},
  booktitle={International Conference on Machine Learning},
  pages={26223--26238},
  year={2022}
}

@misc{2111.04936,
  author = {Zihan Wang and Jialin Lu and Oliver Snow and Martin Ester},
  title = {An Interactive Visualization Tool for Understanding Active Learning},
  year = {2021},
  primaryClass = {cs.LG cs.HC},
  url = {https://arxiv.org/abs/2111.04936}
}

@misc{2201.03947,
  author = {Simon Reichhuber and Sven Tomforde},
  title = {Active Reinforcement Learning -- A Roadmap Towards Curious Classifier Systems for Self-Adaptation},
  year = {2022},
  primaryClass = {cs.LG},
  url = {https://arxiv.org/abs/2201.03947}
}

@misc{2202.00254,
  author = {Shayne Longpre and Julia Reisler and Edward Greg Huang and Yi Lu and Andrew Frank and Nikhil Ramesh and Chris DuBois},
  title = {Active Learning Over Multiple Domains in Natural Language Tasks},
  year = {2022},
  primaryClass = {cs.CL cs.LG},
  url = {https://arxiv.org/abs/2202.00254}
}

@inproceedings{2203.10730,
  title={Semantic Segmentation with Active Semi-Supervised Learning},
  author={Rangnekar, Aneesh and Kanan, Christopher and Hoffman, Matthew},
  booktitle={Proceedings of the IEEE/CVF Winter Conference on Applications of Computer Vision},
  pages={1--10},
  year={2023}
}

@article{2205.13698,
  title={Characterizing the Robustness of Bayesian Adaptive Experimental Designs to Active Learning Bias},
  author={Sloman, Sabina J and Oppenheimer, Daniel M and Broomell, Stephen B and Shalizi, Cosma Rohilla},
  journal={arXiv preprint arXiv:2205.13698},
  year={2022}
}

@article{2207.13339,
  title={ALBench: A Framework for Evaluating Active Learning in Object Detection},
  author={Feng, Zhanpeng and Zhang, Shiliang and Takezoe, Rinyoichi and Hu, Wenze and Chandraker, Manmohan and Li, Li-Jia and Narayanan, Vijay K and Wang, Xiaoyu},
  journal={arXiv preprint arXiv:2207.13339},
  year={2022}
}

@inproceedings{huang2021ymir,
      title={YMIR: A Rapid Data-centric Development Platform for Vision Applications},
      author={Phoenix X. Huang and Wenze Hu and William Brendel and Manmohan Chandraker and Li-Jia Li and Xiaoyu Wang},
      booktitle={Proceedings of the Data-Centric AI Workshop at NeurIPS},
      year={2021},
}

@article{soudi1998optimized,
  title={Optimized distribution protection using binary programming},
  author={Soudi, F and Tomsovic, K},
  journal={IEEE transactions on power delivery},
  volume={13},
  number={1},
  pages={218--224},
  year={1998},
  publisher={IEEE}
}

@article{yanover2006linear,
  title={Linear Programming Relaxations and Belief Propagation--An Empirical Study.},
  author={Yanover, Chen and Meltzer, Talya and Weiss, Yair and Bennett, Kristin P and Parrado-Hern{\'a}ndez, Emilio},
  journal={Journal of Machine Learning Research},
  volume={7},
  number={9},
  year={2006}
}

@article{2210.04675,
  title={A Survey of Methods for Addressing Class Imbalance in Deep-Learning Based Natural Language Processing},
  author={Henning, Sophie and Beluch, William and Fraser, Alexander and Friedrich, Annemarie},
  journal={arXiv preprint arXiv:2210.04675},
  year={2022}
}

@inproceedings{2210.08403,
  title={Semantic Segmentation with Active Semi-Supervised Representation Learning},
  author={Rangnekar, Aneesh and Kanan, Christopher and Hoffman, Matthew},
  booktitle={Proceedings of the British Machine Vision Conference},
  year={2022}
}

@article{2301.01531,
  title={MoBYv2AL: Self-supervised Active Learning for Image Classification},
  author={Caramalau, Razvan and Bhattarai, Binod and Stoyanov, Danail and Kim, Tae-Kyun},
  journal={arXiv preprint arXiv:2301.01531},
  year={2023}
}

@article{2302.05711,
  title={Fair Enough: Standardizing Evaluation and Model Selection for Fairness Research in NLP},
  author={Han, Xudong and Baldwin, Timothy and Cohn, Trevor},
  journal={arXiv preprint arXiv:2302.05711},
  year={2023}
}

@article{2302.14567,
  title={Active Learning with Combinatorial Coverage},
  author={Katragadda, Sai Prathyush and Cody, Tyler and Beling, Peter and Freeman, Laura},
  journal={arXiv preprint arXiv:2302.14567},
  year={2023}
}

@article{2304.05246,
  title={OpenAL: Evaluation and Interpretation of Active Learning Strategies},
  author={Jonas, W. and Abraham, A. and Dreyfus-Schmidt, L.},
  journal={arXiv preprint arXiv:2304.05246},
  year={2023}
}

@article{2305.03900,
  title={Rethinking Class Imbalance in Machine Learning},
  author={Wu, Ou},
  journal={arXiv preprint arXiv:2305.03900},
  year={2023}
}

@article{2306.08001,
  title={A Markovian Formalism for Active Querying},
  author={Ijju, Sid},
  journal={arXiv preprint arXiv:2306.08001},
  year={2023}
}

@article{2306.08954,
  title={Re-Benchmarking Pool-Based Active Learning for Binary Classification},
  author={Lu, Po-Yi and Li, Chun-Liang and Lin, Hsuan-Tien},
  journal={arXiv preprint arXiv:2306.08954},
  year={2023}
}

@article{takezoe2023deep,
  title={Deep active learning for computer vision: Past and future},
  author={Takezoe, Rinyoichi and Liu, Xu and Mao, Shunan and Chen, Marco Tianyu and Feng, Zhanpeng and Zhang, Shiliang and Wang, Xiaoyu and others},
  journal={APSIPA Transactions on Signal and Information Processing},
  volume={12},
  number={1},
  year={2023},
  publisher={Now Publishers, Inc.}
}

@inproceedings{nguyen2019epistemic,
  title={Epistemic uncertainty sampling},
  author={Nguyen, Vu-Linh and Destercke, S{\'e}bastien and H{\"u}llermeier, Eyke},
  booktitle={Discovery Science: 22nd International Conference, DS 2019, Split, Croatia, October 28--30, 2019, Proceedings 22},
  pages={72--86},
  year={2019},
  organization={Springer}
}

@inproceedings{mussmann2018relationship,
  title={On the relationship between data efficiency and error for uncertainty sampling},
  author={Mussmann, Stephen and Liang, Percy},
  booktitle={International Conference on Machine Learning},
  pages={3674--3682},
  year={2018},
  organization={PMLR}
}

@article{hoarau2024evidential,
  title={Evidential uncertainty sampling strategies for active learning},
  author={Hoarau, Arthur and Lemaire, Vincent and Le Gall, Yolande and Dubois, Jean-Christophe and Martin, Arnaud},
  journal={Machine Learning},
  volume={113},
  number={9},
  pages={6453--6474},
  year={2024},
  publisher={Springer}
}

@inproceedings{raj2022convergence,
  title={Convergence of uncertainty sampling for active learning},
  author={Raj, Anant and Bach, Francis},
  booktitle={International conference on machine learning},
  pages={18310--18331},
  year={2022},
  organization={PMLR}
}

@inproceedings{chu2016can,
  title={Can active learning experience be transferred?},
  author={Chu, Hong-Min and Lin, Hsuan-Tien},
  booktitle={2016 IEEE 16th international conference on data mining (ICDM)},
  pages={841--846},
  year={2016},
  organization={IEEE}
}

@inproceedings{kye2023tidal,
  title={TiDAL: Learning training dynamics for active learning},
  author={Kye, Seong Min and Choi, Kwanghee and Byun, Hyeongmin and Chang, Buru},
  booktitle={Proceedings of the IEEE/CVF international conference on computer vision},
  pages={22335--22345},
  year={2023}
}

@article{li2024survey,
  title={A survey on deep active learning: Recent advances and new frontiers},
  author={Li, Dongyuan and Wang, Zhen and Chen, Yankai and Jiang, Renhe and Ding, Weiping and Okumura, Manabu},
  journal={IEEE Transactions on Neural Networks and Learning Systems},
  year={2024},
  publisher={IEEE}
}

@inproceedings{2312.08559,
  title={Fair Active Learning in Low-Data Regimes},
  author={Camilleri, Romain and Wagenmaker, Andrew and Morgenstern, Jamie and Jain, Lalit and Jamieson, Kevin},
  booktitle={Proceedings of the International Conference on Artificial Intelligence and Statistics},
  pages={1585--1593},
  year={2024}
}

@article{2312.09196,
  title={DIRECT: Deep Active Learning under Imbalance and Label Noise},
  author={Nuggehalli, Shyam and Zhang, Jifan and Jain, Lalit and Nowak, Robert},
  journal={arXiv preprint arXiv:2312.09196},
  year={2023}
}

@article{2403.14800,
  title={Deep Active Learning: A Reality Check},
  author={Gashi, Edrina and Deng, Jiankang and Elezi, Ismail},
  journal={arXiv preprint arXiv:2403.14800},
  year={2024}
}

@article{2406.03845,
  title={Open Problem: Active Representation Learning},
  author={Milosevic, Nikola and M{\"u}ller, Gesine and Huisken, Jan and Scherf, Nico},
  journal={arXiv preprint arXiv:2406.03845},
  year={2024}
}

@article{2406.13903,
  title={Generative AI for Enhancing Active Learning in Education: A Comparative Study of GPT-3.5 and GPT-4 in Crafting Customized Test Questions},
  author={Rouzegar, Hamdireza and Makrehchi, Masoud},
  journal={arXiv preprint arXiv:2406.13903},
  year={2024}
}

@article{2407.18745,
  title={FairAIED: Navigating Fairness, Bias, and Ethics in Educational AI Applications},
  author={Chinta, Sribala Vidyadhari and Wang, Zichong and Yin, Zhipeng and Hoang, Nhat and Gonzalez, Matthew and Quy, Tai Le and Zhang, Wenbin},
  journal={arXiv preprint arXiv:2407.18745},
  year={2024}
}

@article{2408.00426,
  title={A Cross-Domain Benchmark for Active Learning},
  author={Werner, Thorben and Burchert, Johannes and Stubbemann, Maximilian and Schmidt-Thieme, Lars},
  journal={arXiv preprint arXiv:2408.00426},
  year={2024}
}

@article{2408.13690,
  title={Understanding Uncertainty-based Active Learning Under Model Mismatch},
  author={Rahmati, Amir Hossein and Fan, Mingzhou and Zhou, Ruida and Urban, Nathan M and Yoon, Byung-Jun and Qian, Xiaoning},
  journal={arXiv preprint arXiv:2408.13690},
  year={2024}
}

@article{2410.02145,
  title={Active Learning of Deep Neural Networks via Gradient-Free Cutting Planes},
  author={Zhang, Erica and Zhang, Fangzhao and Pilanci, Mert},
  journal={arXiv preprint arXiv:2410.02145},
  year={2024}
}

@article{2412.02310,
  title={Active Learning via Classifier Impact and Greedy Selection for Interactive Image Retrieval},
  author={Bar, Leah and Lerner, Boaz and Darshan, Nir and Ben-Ari, Rami},
  journal={arXiv preprint arXiv:2412.02310},
  year={2024}
}

@article{2412.11388,
  title={INTERACT: Enabling Interactive, Question-Driven Learning in Large Language Models},
  author={Kendapadi, Aum and Zaman, Kerem and Menon, Rakesh R and Srivastava, Shashank},
  journal={arXiv preprint arXiv:2412.11388},
  year={2024}
}

@incollection{lewis1994heterogeneous,
  title={Heterogeneous uncertainty sampling for supervised learning},
  author={Lewis, David D and Catlett, Jason},
  booktitle={Machine learning proceedings 1994},
  pages={148--156},
  year={1994},
  publisher={Elsevier}
}

@article{luo2013latent,
  title={Latent structured active learning},
  author={Luo, Wenjie and Schwing, Alex and Urtasun, Raquel},
  journal={Advances in neural information processing systems},
  volume={26},
  year={2013}
}

@inproceedings{lewis1995sequential,
  title={A sequential algorithm for training text classifiers: Corrigendum and additional data},
  author={Lewis, David D},
  booktitle={Acm Sigir Forum},
  volume={29},
  pages={13--19},
  year={1995},
  organization={ACM New York, NY, USA}
}

@inproceedings{settles2008analysis,
  title={An analysis of active learning strategies for sequence labeling tasks},
  author={Settles, Burr and Craven, Mark},
  booktitle={proceedings of the 2008 conference on empirical methods in natural language processing},
  pages={1070--1079},
  year={2008}
}

@article{vijayanarasimhan2014large,
  title={Large-scale live active learning: Training object detectors with crawled data and crowds},
  author={Vijayanarasimhan, Sudheendra and Grauman, Kristen},
  journal={International journal of computer vision},
  volume={108},
  pages={97--114},
  year={2014},
  publisher={Springer}
}

@inproceedings{iglesias2011combining,
  title={Combining generative and discriminative models for semantic segmentation of CT scans via active learning},
  author={Iglesias, Juan Eugenio and Konukoglu, Ender and Montillo, Albert and Tu, Zhuowen and Criminisi, Antonio},
  booktitle={Biennial international conference on information processing in medical imaging},
  pages={25--36},
  year={2011},
  organization={Springer}
}

@InProceedings{10.1007/978-3-642-41398-8_16,
author="Evans, Lewis P. G.
and Adams, Niall M.
and Anagnostopoulos, Christoforos",
editor="Tucker, Allan
and H{\"o}ppner, Frank
and Siebes, Arno
and Swift, Stephen",
title="When Does Active Learning Work?",
booktitle="Advances in Intelligent Data Analysis XII",
year="2013",
publisher="Springer Berlin Heidelberg",
address="Berlin, Heidelberg",
pages="174--185",
isbn="978-3-642-41398-8"
}

@article{ding2023active,
  title={Active learning in physics: From 101, to progress, and perspective},
  author={Ding, Yongcheng and Mart{\'\i}n-Guerrero, Jos{\'e} D and Vives-Gilabert, Yolanda and Chen, Xi},
  journal={Advanced Quantum Technologies},
  pages={2300208},
  year={2023},
  publisher={Wiley Online Library}
}

@inproceedings{abraham2020rebuilding,
  title={Rebuilding trust in active learning with actionable metrics},
  author={Abraham, Alexandre and Dreyfus-Schmidt, L{\'e}o},
  booktitle={2020 International Conference on Data Mining Workshops (ICDMW)},
  pages={836--843},
  year={2020},
  organization={IEEE}
}

@article{zhang2022survey,
  title={A survey of active learning for natural language processing},
  author={Zhang, Zhisong and Strubell, Emma and Hovy, Eduard},
  journal={arXiv preprint arXiv:2210.10109},
  year={2022}
}

@inproceedings{jukic-snajder-2023-smooth,
    title = "Smooth Sailing: Improving Active Learning for Pre-trained Language Models with Representation Smoothness Analysis",
    author = "Juki{\'c}, Josip  and
      Snajder, Jan",
    editor = "Breitholtz, Ellen  and
      Lappin, Shalom  and
      Loaiciga, Sharid  and
      Ilinykh, Nikolai  and
      Dobnik, Simon",
    booktitle = "Proceedings of the 2023 CLASP Conference on Learning with Small Data (LSD)",
    month = sep,
    year = "2023",
    address = "Gothenburg, Sweden",
    publisher = "Association for Computational Linguistics",
    url = "https://aclanthology.org/2023.clasp-1.2/",
    pages = "11--24"
}

@article{jiang2024actively,
  title={Actively learning to learn causal relationships},
  author={Jiang, Chentian and Lucas, Christopher G},
  journal={Computational Brain \& Behavior},
  volume={7},
  number={1},
  pages={80--105},
  year={2024},
  publisher={Springer}
}

@inproceedings{misra2018learning,
  title={Learning by asking questions},
  author={Misra, Ishan and Girshick, Ross and Fergus, Rob and Hebert, Martial and Gupta, Abhinav and Van Der Maaten, Laurens},
  booktitle={Proceedings of the IEEE Conference on Computer Vision and Pattern Recognition},
  pages={11--20},
  year={2018}
}

@inproceedings{li2022more,
  title={When more data lead us astray: Active data acquisition in the presence of label bias},
  author={Li, Yunyi and De-Arteaga, Maria and Saar-Tsechansky, Maytal},
  booktitle={Proceedings of the AAAI Conference on Human Computation and Crowdsourcing},
  volume={10},
  pages={133--146},
  year={2022}
}

@article{10.14778/3641204.3641207,
author = {Tae, Ki Hyun and Zhang, Hantian and Park, Jaeyoung and Rong, Kexin and Whang, Steven Euijong},
title = {Falcon: Fair Active Learning Using Multi-Armed Bandits},
year = {2024},
issue_date = {January 2024},
publisher = {VLDB Endowment},
volume = {17},
number = {5},
issn = {2150-8097},
url = {https://doi.org/10.14778/3641204.3641207},
doi = {10.14778/3641204.3641207},
journal = {Proc. VLDB Endow.},
month = jan,
pages = {952–965},
numpages = {14}
}

@inproceedings{bengar2022class,
  title={Class-balanced active learning for image classification},
  author={Bengar, Javad Zolfaghari and van de Weijer, Joost and Fuentes, Laura Lopez and Raducanu, Bogdan},
  booktitle={Proceedings of the IEEE/CVF winter conference on applications of computer vision},
  pages={1536--1545},
  year={2022}
}

@inproceedings{zhang2022galaxy,
  title={Galaxy: Graph-based active learning at the extreme},
  author={Zhang, Jifan and Katz-Samuels, Julian and Nowak, Robert},
  booktitle={International Conference on Machine Learning},
  pages={26223--26238},
  year={2022},
  organization={PMLR}
}

@inproceedings{gilhuber2023overcome,
  title={How to overcome confirmation bias in semi-supervised image classification by active learning},
  author={Gilhuber, Sandra and Hvingelby, Rasmus and Fok, Mang Ling Ada and Seidl, Thomas},
  booktitle={Joint European Conference on Machine Learning and Knowledge Discovery in Databases},
  pages={330--347},
  year={2023},
  organization={Springer}
}

@inproceedings{zhao2021active,
  title={Active learning under label shift},
  author={Zhao, Eric and Liu, Anqi and Anandkumar, Animashree and Yue, Yisong},
  booktitle={International Conference on artificial intelligence and statistics},
  pages={3412--3420},
  year={2021},
  organization={PMLR}
}

@article{aaboud2018search,
  title={Search for the standard model Higgs boson produced in association with top quarks and decaying into abb{\={}} pair in pp collisions at s= 13 TeV with the ATLAS detector},
  author={Aaboud, Morad and Aad, Georges and Abbott, Brad and Abdinov, Ovsat and Abeloos, Baptiste and Abidi, Syed Haider and AbouZeid, OS and Abraham, Nadine L and Abramowicz, Halina and Abreu, Henso and others},
  journal={Physical Review D},
  volume={97},
  number={7},
  pages={072016},
  year={2018},
  publisher={APS}
}

@article{grimm2015self,
  title={Self-dual tensors and partial supersymmetry breaking in five dimensions},
  author={Grimm, Thomas W and Kapfer, Andreas},
  journal={Journal of High Energy Physics},
  volume={2015},
  number={3},
  pages={1--31},
  year={2015},
  publisher={Springer}
}

@inproceedings{iyyer2017amazing,
  title={The amazing mysteries of the gutter: Drawing inferences between panels in comic book narratives},
  author={Iyyer, Mohit and Manjunatha, Varun and Guha, Anupam and Vyas, Yogarshi and Boyd-Graber, Jordan and Daume, Hal and Davis, Larry S},
  booktitle={Proceedings of the IEEE Conference on Computer Vision and Pattern recognition},
  pages={7186--7195},
  year={2017}
}

@article{tuia2012learning,
  title={Learning user's confidence for active learning},
  author={Tuia, Devis and Munoz-Mari, Jordi},
  journal={IEEE Transactions on Geoscience and Remote Sensing},
  volume={51},
  number={2},
  pages={872--880},
  year={2012},
  publisher={IEEE}
}

@article{qu2025recent,
  title={Recent advances of continual learning in computer vision: An overview},
  author={Qu, Haoxuan and Rahmani, Hossein and Xu, Li and Williams, Bryan and Liu, Jun},
  journal={IET Computer Vision},
  volume={19},
  number={1},
  pages={e70013},
  year={2025},
  publisher={Wiley Online Library}
}

@inproceedings{kyrkou2020imitation,
  title={Imitation-based active camera control with deep convolutional neural network},
  author={Kyrkou, Christos},
  booktitle={2020 IEEE 4th International Conference on Image Processing, Applications and Systems (IPAS)},
  pages={168--173},
  year={2020},
  organization={IEEE}
}

@inproceedings{jeon2017active,
  title={Active convolution: Learning the shape of convolution for image classification},
  author={Jeon, Yunho and Kim, Junmo},
  booktitle={Proceedings of the IEEE conference on computer vision and pattern recognition},
  pages={4201--4209},
  year={2017}
}

@inproceedings{parvaneh2022active,
  title={Active learning by feature mixing},
  author={Parvaneh, Amin and Abbasnejad, Ehsan and Teney, Damien and Haffari, Gholamreza Reza and Van Den Hengel, Anton and Shi, Javen Qinfeng},
  booktitle={Proceedings of the IEEE/CVF conference on computer vision and pattern recognition},
  pages={12237--12246},
  year={2022}
}

@inproceedings{kohl2024scoping,
  title={Scoping review of active learning strategies and their evaluation environments for entity recognition tasks},
  author={Kohl, Philipp and Kr{\"a}mer, Yoka and Fohry, Claudia and Kraft, Bodo},
  booktitle={International Conference on Deep Learning Theory and Applications},
  pages={84--106},
  year={2024},
  organization={Springer}
}

@inproceedings{jiang2021application,
  title={The Application of Active Query K-Means in Text Classification},
  author={Jiang, Yukun},
  booktitle={2021 3rd International Conference on Natural Language Processing (ICNLP)},
  pages={20--25},
  year={2021},
  organization={IEEE}
}

@inproceedings{bang2024active,
  title={Active prompt learning in vision language models},
  author={Bang, Jihwan and Ahn, Sumyeong and Lee, Jae-Gil},
  booktitle={Proceedings of the IEEE/CVF Conference on Computer Vision and Pattern Recognition},
  pages={27004--27014},
  year={2024}
}

@incollection{virtanen2022product,
  title={On the product formula for Toeplitz and related operators},
  author={Virtanen, Jani A},
  booktitle={Toeplitz Operators and Random Matrices: In Memory of Harold Widom},
  pages={605--616},
  year={2022},
  publisher={Springer}
}

@article{humeniuk2023ambiegen,
  title={Ambiegen: A search-based framework for autonomous systems testing},
  author={Humeniuk, Dmytro and Khomh, Foutse and Antoniol, Giuliano},
  journal={Science of Computer Programming},
  volume={230},
  pages={102990},
  year={2023},
  publisher={Elsevier}
}

@inproceedings{su2020active,
  title={Active adversarial domain adaptation},
  author={Su, Jong-Chyi and Tsai, Yi-Hsuan and Sohn, Kihyuk and Liu, Buyu and Maji, Subhransu and Chandraker, Manmohan},
  booktitle={Proceedings of the IEEE/CVF winter conference on applications of computer vision},
  pages={739--748},
  year={2020}
}

@inproceedings{huang2023divide,
  title={Divide and adapt: Active domain adaptation via customized learning},
  author={Huang, Duojun and Li, Jichang and Chen, Weikai and Huang, Junshi and Chai, Zhenhua and Li, Guanbin},
  booktitle={Proceedings of the IEEE/CVF conference on computer vision and pattern recognition},
  pages={7651--7660},
  year={2023}
}

@inproceedings{xu2019larger,
  title={Larger norm more transferable: An adaptive feature norm approach for unsupervised domain adaptation},
  author={Xu, Ruijia and Li, Guanbin and Yang, Jihan and Lin, Liang},
  booktitle={Proceedings of the IEEE/CVF international conference on computer vision},
  pages={1426--1435},
  year={2019}
}

@article{zhou2021discriminative,
  title={Discriminative active learning for domain adaptation},
  author={Zhou, Fan and Shui, Changjian and Yang, Shichun and Huang, Bincheng and Wang, Boyu and Chaib-draa, Brahim},
  journal={Knowledge-Based Systems},
  volume={222},
  pages={106986},
  year={2021},
  publisher={Elsevier}
}

@article{chen2020domain,
  title={Domain adversarial reinforcement learning for partial domain adaptation},
  author={Chen, Jin and Wu, Xinxiao and Duan, Lixin and Gao, Shenghua},
  journal={IEEE Transactions on Neural Networks and Learning Systems},
  volume={33},
  number={2},
  pages={539--553},
  year={2020},
  publisher={IEEE}
}

@article{dakic2017macroscopic,
  title={Macroscopic superpositions as quantum ground states},
  author={Daki{\'c}, Borivoje and Radonji{\'c}, Milan},
  journal={Physical Review Letters},
  volume={119},
  number={9},
  pages={090401},
  year={2017},
  publisher={APS}
}

@article{baldassi2016unreasonable,
  title={Unreasonable effectiveness of learning neural networks: From accessible states and robust ensembles to basic algorithmic schemes},
  author={Baldassi, Carlo and Borgs, Christian and Chayes, Jennifer T and Ingrosso, Alessandro and Lucibello, Carlo and Saglietti, Luca and Zecchina, Riccardo},
  journal={Proceedings of the National Academy of Sciences},
  volume={113},
  number={48},
  pages={E7655--E7662},
  year={2016},
  publisher={National Academy of Sciences}
}

@inproceedings{tan2018survey,
  title={A survey on deep transfer learning},
  author={Tan, Chuanqi and Sun, Fuchun and Kong, Tao and Zhang, Wenchang and Yang, Chao and Liu, Chunfang},
  booktitle={Artificial Neural Networks and Machine Learning--ICANN 2018: 27th International Conference on Artificial Neural Networks, Rhodes, Greece, October 4-7, 2018, Proceedings, Part III 27},
  pages={270--279},
  year={2018},
  organization={Springer}
}

@inproceedings{xie2020deal,
  title={Deal: Difficulty-aware active learning for semantic segmentation},
  author={Xie, Shuai and Feng, Zunlei and Chen, Ying and Sun, Songtao and Ma, Chao and Song, Mingli},
  booktitle={Proceedings of the Asian conference on computer vision},
  year={2020}
}

@article{goupilleau2021active,
  title={Active learning for object detection in high-resolution satellite images},
  author={Goupilleau, Alex and Ceillier, Tugdual and Corbineau, Marie-Caroline},
  journal={arXiv preprint arXiv:2101.02480},
  year={2021}
}

@inproceedings{wang2023alwod,
  title={Alwod: Active learning for weakly-supervised object detection},
  author={Wang, Yuting and Ilic, Velibor and Li, Jiatong and Kisa{\v{c}}anin, Branislav and Pavlovic, Vladimir},
  booktitle={Proceedings of the IEEE/CVF international conference on computer vision},
  pages={6459--6469},
  year={2023}
}

@inproceedings{probst2022evaluating,
  title={Evaluating Zero-Cost Active Learning for Object Detection},
  author={Probst, Dominik and Raza, Hasnain and Rodner, Erik},
  booktitle={International Conference on Software Engineering and Formal Methods},
  pages={38--47},
  year={2022},
  organization={Springer}
}

@inproceedings{mittal2023best,
  title={Best practices in active learning for semantic segmentation},
  author={Mittal, Sudhanshu and Niemeijer, Joshua and Sch{\"a}fer, J{\"o}rg P and Brox, Thomas},
  booktitle={DAGM German Conference on Pattern Recognition},
  pages={427--442},
  year={2023},
  organization={Springer}
}

@inproceedings{beatty2019use,
  title={The use of unlabeled data versus labeled data for stopping active learning for text classification},
  author={Beatty, Garrett and Kochis, Ethan and Bloodgood, Michael},
  booktitle={2019 IEEE 13th International Conference on Semantic Computing (ICSC)},
  pages={287--294},
  year={2019},
  organization={IEEE}
}

@inproceedings{choi2021vab,
  title={Vab-al: Incorporating class imbalance and difficulty with variational bayes for active learning},
  author={Choi, Jongwon and Yi, Kwang Moo and Kim, Jihoon and Choo, Jinho and Kim, Byoungjip and Chang, Jinyeop and Gwon, Youngjune and Chang, Hyung Jin},
  booktitle={Proceedings of the IEEE/CVF conference on computer vision and pattern recognition},
  pages={6749--6758},
  year={2021}
}

@article{unal2023discwise,
  title={Discwise active learning for lidar semantic segmentation},
  author={Unal, Ozan and Dai, Dengxin and Unal, Ali Tamer and Van Gool, Luc},
  journal={IEEE Robotics and Automation Letters},
  volume={8},
  number={11},
  pages={7671--7678},
  year={2023},
  publisher={IEEE}
}

@InProceedings{Li_2019_CVPR,
author = {Li, Yi and Vasconcelos, Nuno},
title = {REPAIR: Removing Representation Bias by Dataset Resampling},
booktitle = {Proceedings of the IEEE/CVF Conference on Computer Vision and Pattern Recognition (CVPR)},
month = {June},
year = {2019}
}
